\documentclass[journal,twocolumn]{IEEEtran}

\usepackage{cite}
\usepackage{graphicx}
\usepackage{amsmath}
\usepackage{wrapfig}
\usepackage{array}
\usepackage[caption=false,font=footnotesize]{subfig}
\usepackage{url}
\graphicspath{{./fig/}}
\usepackage[T1]{fontenc}
\usepackage[usenames, dvipsnames]{color}
\usepackage{courier}
\usepackage{hyperref}
\usepackage{float}
\usepackage{subfig}
\usepackage{breakurl}
\usepackage[normalem]{ulem}
\usepackage{listings}
\usepackage{verbatim}
\definecolor{mypink1}{rgb}{0.858, 0.188, 0.478}
\definecolor{mypink2}{RGB}{219, 48, 122}
\definecolor{mypink3}{cmyk}{0, 0.7808, 0.4429, 0.1412}
\definecolor{mygray}{gray}{0.3}

\lstdefinestyle{BashInputStyle}{
  language=bash,
  basicstyle=\footnotesize\ttfamily,
  frame=tb,
  columns=fullflexible,
  linewidth=0.9\linewidth,
}

\begin{document}

\title{Managing a Fleet of Autonomous Mobile Robots (AMR) using Cloud Robotics Platform}

\author{Aniruddha Singhal, Nishant Kejriwal, Prasun Pallav, Soumyadeep Choudhury, Rajesh Sinha and Swagat Kumar 
\thanks{Email IDs: \texttt{\{aniruddha.singhal, nishant.kejriwal, prasun.pallav, soumyadeep.choudhury, rajesh.sinha, swagat.kumar\}@tcs.com}}
\thanks{The authors are with TCS Research, Tata Consultancy Services, New Delhi, India 201309.} }

\maketitle

\begin{abstract}
  In this paper, we provide details of implementing a system for
  managing a fleet of autonomous mobile robots (AMR) operating in a
  factory or a warehouse premise. While the robots are themselves
  autonomous in its motion and obstacle avoidance capability, the
  target destination for each robot is provided by a global planner.
  The global planner and the ground vehicles (robots) constitute a
  multi agent system (MAS) which communicate with each other over a
  wireless network. Three different approaches are explored for
  implementation. The first two approaches make use of the distributed
  computing based Networked Robotics architecture and communication
  framework of Robot Operating System (ROS) itself while the third
  approach uses Rapyuta Cloud Robotics framework for this
  implementation. The comparative performance of these approaches are
  analyzed through simulation as well as real world experiment with
  actual robots. These analyses provide an in-depth understanding of
  the inner working of the Cloud Robotics Platform in contrast to the
  usual ROS framework. The insight gained through this exercise will
  be valuable for students as well as   practicing engineers
  interested in implementing similar systems else where.  In the
  process, we also identify few critical limitations of the current
  Rapyuta platform and provide suggestions to overcome them.
\end{abstract}



\IEEEpeerreviewmaketitle

\section{Introduction}\label{sec:int}

The last couple of decades have witnessed a steady rise in robot-based
industrial automation. These industrial robots are comparatively
inexpensive and are capable of carrying out repeated tasks at high
speed and great accuracy and hence, are widely deployed in the
industries of mass production. In spite of this, the robotic
automation has remained confined only to big industries who can pay
for elaborate assembly lines built around these robots to compensate
for their lack of intelligence. In addition, this involves writing
and testing extensive programs to take into account all possible cases
that a robot might encounter during its operation. In short, the
current robot-based industrial automation  requires huge investment
both in terms of capital and time, making it unaffordable to small and
medium enterprises. This scenario is poised to change with the rise of
service robots \cite{wettergreen2016field}
\cite{amirat2016assistance}, which unlike their industrial
counterparts, can work in unstructured environments while
learning and adapting to 
\begin{figure}[!t]
  \centering
  \includegraphics[width=3.5in]{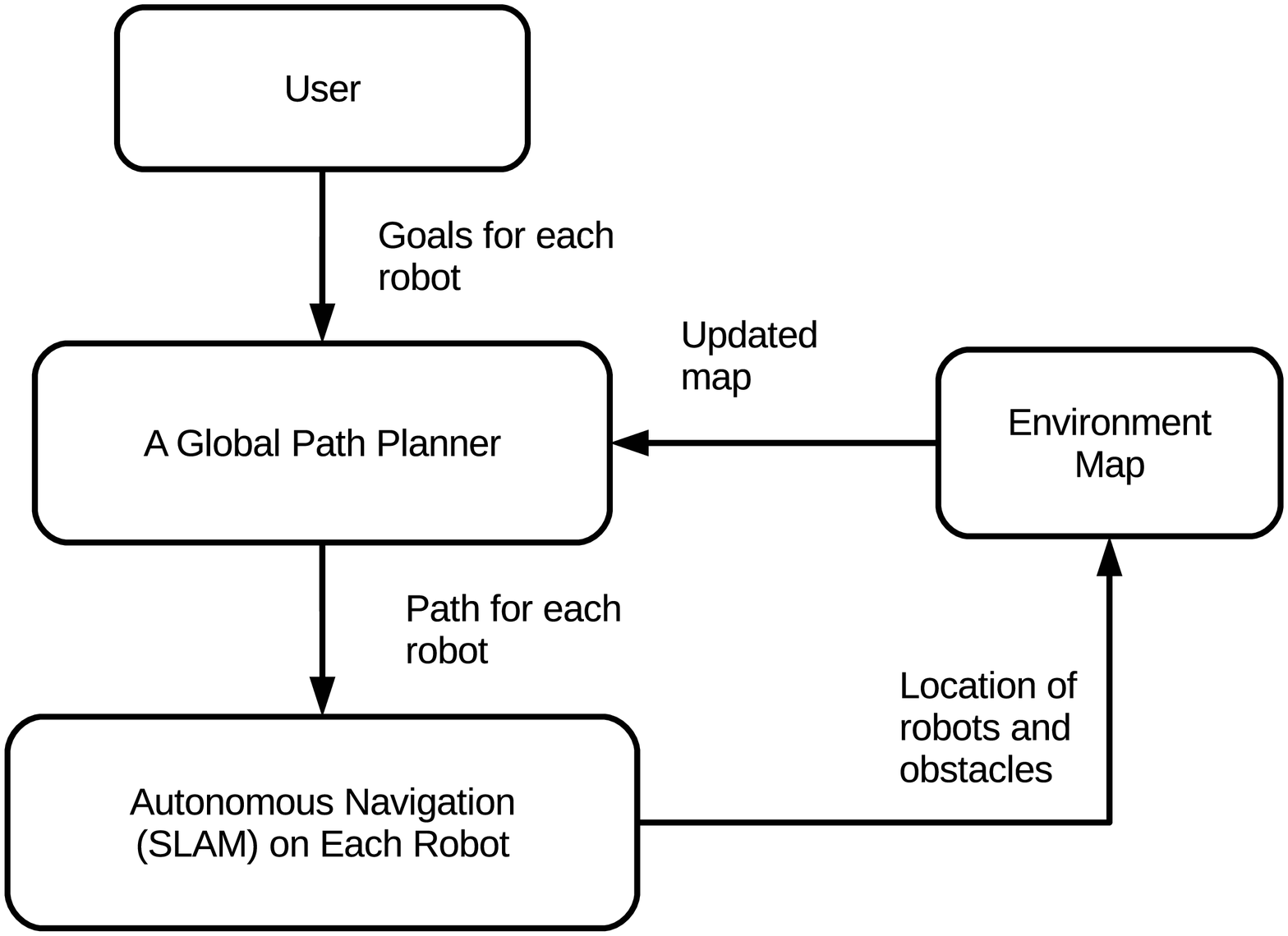}
  \caption{Block diagram of a simplified fleet management system for
  autonomous vehicles.}
  \label{fig:fms_bd}
\end{figure}changes around them. These robots are
designed to be safe and can work collaboratively with humans in close
proximity without any protective fencing. These robots could be
programmed very easily and intuitively through demonstrations by
operators themselves. This gives rise to a field known as teaching by
demonstration paradigm \cite{argall2009survey} which can be used for
changing the robot behavior on the fly. Similarly, these robots will
exhibit higher level of intelligence in taking autonomous decisions
based on sensory perception.  

According to International Federation of Robotics (IFR)
\cite{ifrserv}, service robotics is going to drive the growth in
robotic industry in the coming decade. This growth will be partly due
to the increased adoption of robots in industries as well as domestic
environments.  Cloud Robotics \cite{mohanarajah2015rapyuta}
\cite{kehoe2015survey} will play a significant role in the growth of
service robotics by augmenting the robot capabilities while reducing
the per unit costs of each robot.  This will become possible as the
robots can off-load computationally intensive tasks on to the cloud
for processing, can collaborate with other robots and humans over
network, can learn new skills instantly from internet. Cloud Robotics
can be used for providing ``Robotics-as-a-Service'' based solutions
where robots could be dynamically combined to give support to specific
applications \cite{koken2015evolution}. One such application that is
being considered in this paper is a vehicle fleet management system
for warehouse and factory shop floors.

A vehicle fleet management system comprises various software and
hardware components which facilitates optimum utilization of vehicles
in meeting pre-defined goals. One such example is the use of Kiva
mobile robots \cite{wurman2008coordinating} for moving goods within
Amazon fulfillment centers \cite{amazon}. These autonomous ground
vehicles (AGVs) are programmed to move autonomously along predefined
tracks.  However, the schedule and routes are provided by a
centralized planner which also carries out resource allocation and
manages job assignment to individual robots. Such a system also
includes effective modules that facilitates efficient collaboration
between machines and robots \cite{rosenfeld2016human}. 

In this paper, we are looking into a simpler version of this fleet
management system where a group of autonomous vehicles are required to
follow desired paths provided by a global path planner as shown in
Figure \ref{fig:fms_bd}. This figure shows the essential components
required for implementing such a fleet management system. The current
location of robots as well as new obstacles detected on the way are
used to update the environment map which, in turn, is used by the
global planner to create new paths for the robots.  The user or the
operator provides the goals or destinations for each robot in this
case. However, such goals may also come from an ERP (Enterprise
Resource Planning) system in an industrial setting. The autonomy of
each robot is governed by the navigation module that implements SLAM
(Simultaneous localization and mapping) \cite{thrun2005probabilistic}
as well as obstacle avoidance capabilities. Unlike the existing
systems that focus on system integration involving various software
and hardware components \cite{wellman2012fleet}
\cite{coffee2003vehicle} \cite{schnell2014fleet}, we are particularly
interested in exploring various software frameworks like ROS
\cite{ros} and Rapyuta \cite{mohanarajah2015rapyuta} for implementing
such systems.  To be specific, we provide details of three
implementation in this paper. First two make use of the distributed
control and communication framework of Robot Operating System (ROS)
\cite{ros} and the last implementation uses Rapyuta cloud robotics
engine \cite{mohanarajah2015rapyuta}. A comparative analysis of these
approaches are carried out which provides an understanding of
underlying challenges, which if addressed, may increase the usability
of the platform. The working of these implementations are demonstrated
through several simulation as well as real world experiments. 

In short, the contributions made in this paper could be summarized as
follows: (1) We provide three different implementations of a fleet
management system for autonomous ground vehicles using ROS and Rapyuta
platforms. This includes single-master based ROS system, multi-master
based ROS system and Rapyuta-based cloud robotics system. (2) The
working details of these implementations are provided for both
simulation as well as actual experiments which could serve as
operation manual for students, researchers and practicing engineers
who would like to implement similar systems in other domains. (3)
Through rigorous comparative performance analysis, we identify the
critical limitations of existing cloud robotics platform which, if
solved, will improve the usability of these platforms.  

The rest of this paper is organized as follows. An overview of related
work is provided in the next section. The three approaches of
implementing fleet management system is described in Section
\ref{sec:meth}. The comparative performance analysis of these
systems for simulation and actual experiments are provided in Sections
\ref{sec:sim} and \ref{sec:expt} respectively. The limitations of the
current implementation which provides direction for future work is
discussed in Section \ref{sec:limit} followed by conclusion in
Section \ref{sec:conc}.

\section{Related Work} \label{sec:relw}

In this section we provide a brief overview of several related work.
This will also serve as a background material for various core
concepts that will be repeatedly referred in the rest of this
paper.

\subsection{Robot Operating System} \label{sec:ros}

Robot Operating System (ROS) \cite{quigley2009ros} is a software
framework for managing and controlling multiple robots. It uses a
peer-to-peer topology for communication between robot processes,
supports multiple programming languages and provides tools for robot
software development.  Readers can refer to online wiki \cite{ros} to
know about ROS in detail. For the sake of completion, some of the
common concepts which will be used frequently are listed below for the
sake of completeness.  

 (1) Nodes are \texttt{ROS} processes that perform
  computation. They can communicate with each other by passing
  messages. 
 (2) Topics are medium over which nodes exchange messages.
  They provide a link between two nodes. A topic is channel for
  anonymous communication. Multiple nodes can publish/subscribe to a
  given topic.  
 (3) Subscriber is a node which listens to the messages that
  are published to a topic.
(4) Publisher is a node which writes to a topic from which
  other nodes can subscribe.
(5) roscore is a set of nodes which are necessary for ROS
  environment to work. roscore starts a ROS master node, ROS permanent
  server and a node where logs are published.
 (6) AMCL  (Adaptive Monte Carlo Localization)\cite{fox1999monte} is an inbuilt
  package in ROS that is used by the robots to localizes themselves in
  the map.
(7) TF is a package that lets the user keep track of
  multiple coordinate frames over time. \textbf{TF} maintains the relationship
  between coordinate frames in a tree structure buffered in time, and
  lets the user transform points, vectors, etc., between any two
  coordinate frames at any desired point in time.
(8) GMapping \cite{grisetti2007improved} package provides laser-based SLAM
  (Simultaneous Localization and Mapping) capability. It runs as a ROS node called
  \texttt{slam\_gmapping}. This node can be used for creating 2-D occupancy
  grid map of the environment from laser and pose data collected by a mobile robot.

\subsection{Cloud Robotics Platform: Rapyuta} \label{sec:cr}

Rapyuta \cite{mohanarajah2015rapyuta} is an
open-source cloud robotics framework. It provides an elastic computing
model which dynamically allocates secure computing environment for
robots. In this way it helps in solving the problem of unavailability
of high computing power on robots. The Rapyuta framework is based on
clone-based model \cite{hu2012cloud} where each robot connected to the
cloud has a system level clone on the cloud itself which allows them
to offload heavy computation into the cloud.  These clones are tightly
interconnected with high bandwidth making it suitable for multi-robot
deployment. In addition, Rapyuta provides access to libraries of
images, maps otherwise known as RoboEarth knowledge repository
\cite{tenorth2012roboearth} \cite{zweigle2009roboearth} and, provides
framework that facilitates collaborative robot learning and human
computation \cite{kehoe2015survey}. A number of applications have been
reported in literature that demonstrate the applicability and
usefulness of the platform. This includes collaborative mapping
\cite{hu2012cloud} \cite{mohanarajah2015cloud}, robot-grasping
\cite{kehoe2013cloud}, tele-presence \cite{ng2015cloud} and ubiquitous
manufacturing \cite{wang2016ubiquitous}.  Readers are also referred to
\cite{mohanarajah2015cloud} \cite{goldberg2013cloud}
\cite{goldberg2013cloud} for a comparative study on several other
cloud robotics platforms reported in the literature. While a
cloud-based system offers several advantages, it also poses several
challenges which if solved can greatly enhance the usability of such
platforms. Some of these challenges include network latency, data
interaction and security \cite{wan2016cloud}.

Also a slightly related work is done by Turnbull et al in which they
have made a system to detect position of robots through a camera
placed on ceiling and control their motion so that they don't collide.
They have exploited the large computation power provided by the
cloud.\cite{turnbull2013cloud}. A collision avoidance and path
planning system which works on individual robots also exist
\cite{hennes2012multi}. They have used common ROS topic for inter
robot communication and AMCL for localization.

\subsection{Fleet Management System} \label{sec:fms}

A fleet management system \cite{wellman2012fleet}
\cite{coffee2004vehicle} \cite{schnell2014fleet}
\cite{thong2007intelligent} primarily concerns itself with managing a
group of vehicles to meet the goals and objectives obtained from an
enterprise computer system. While most of the existing system focus on
integrating various software and hardware components to ensure
efficient utilization of resources, there has been very few efforts at
generalizing the underlying architecture to make it more flexible and
generic.  Authors in \cite{turnbull2013cloud} do propose to use a
cloud infrastructure to implement formation control of a multi-robot
system by using an external camera system for detecting and tracking
individual robots. While a cloud infrastructure is used for image
processing, it does not use a generic framework like Rapyuta.  

In this paper, we primarily implement a simplified fleet management
system using Rapyuta cloud robotics engine. The implementation is
carried out through simulation as well as physical experiments using
actual robots. The purpose of this work is to provide an insight into
the working of the cloud robotics framework as well as identifying the
limitations of current architecture. We also attempt to offer
suggestions for overcoming these limitations and thereby improving the
usability of the Rapyuta cloud robotics framework.
The details of implementations for fleet management
  system is described next in this paper.

\section{The Methods} \label{sec:meth} 

In this section, we provide details of our
implementation of a simplified fleet management system as shown in
Figure \ref{fig:fms_bd}. It primarily consists of
  four modules: (1) a user, an operator or an ERP system that
  provides goals or target destination for each robot, (2) a global
  planner that computes the path to be taken by each robot based on
  the current state of the environment (3) Autonomous Mobile Robots
  (AMR) having capability for autonomous navigation and obstacle
  avoidance; and (4) an environment map which could be updated with
  the information of new obstacles detected by the robots. The user
  is also free to update the availability of routes for any robot by
creating obstacles in the environment map. 

The above fleet management system is implemented using three methods:
(1) single-master system, (2) multi-master system and (3) Cloud
Robotics platform. The first two methods make use of the distributed
computing and communication architecture of Robot Operating System
(ROS) \cite{quigley2009ros} while the last methods uses Rapyuta cloud
robotics framework \cite{mohanarajah2015rapyuta}. The details of each
implementation and their respective pros and cons are presented next
in this section.

\subsection{Single Master System} \label{sec:sms} 

\begin{figure}[!t] 
  \centering 
  \includegraphics[width=\columnwidth]{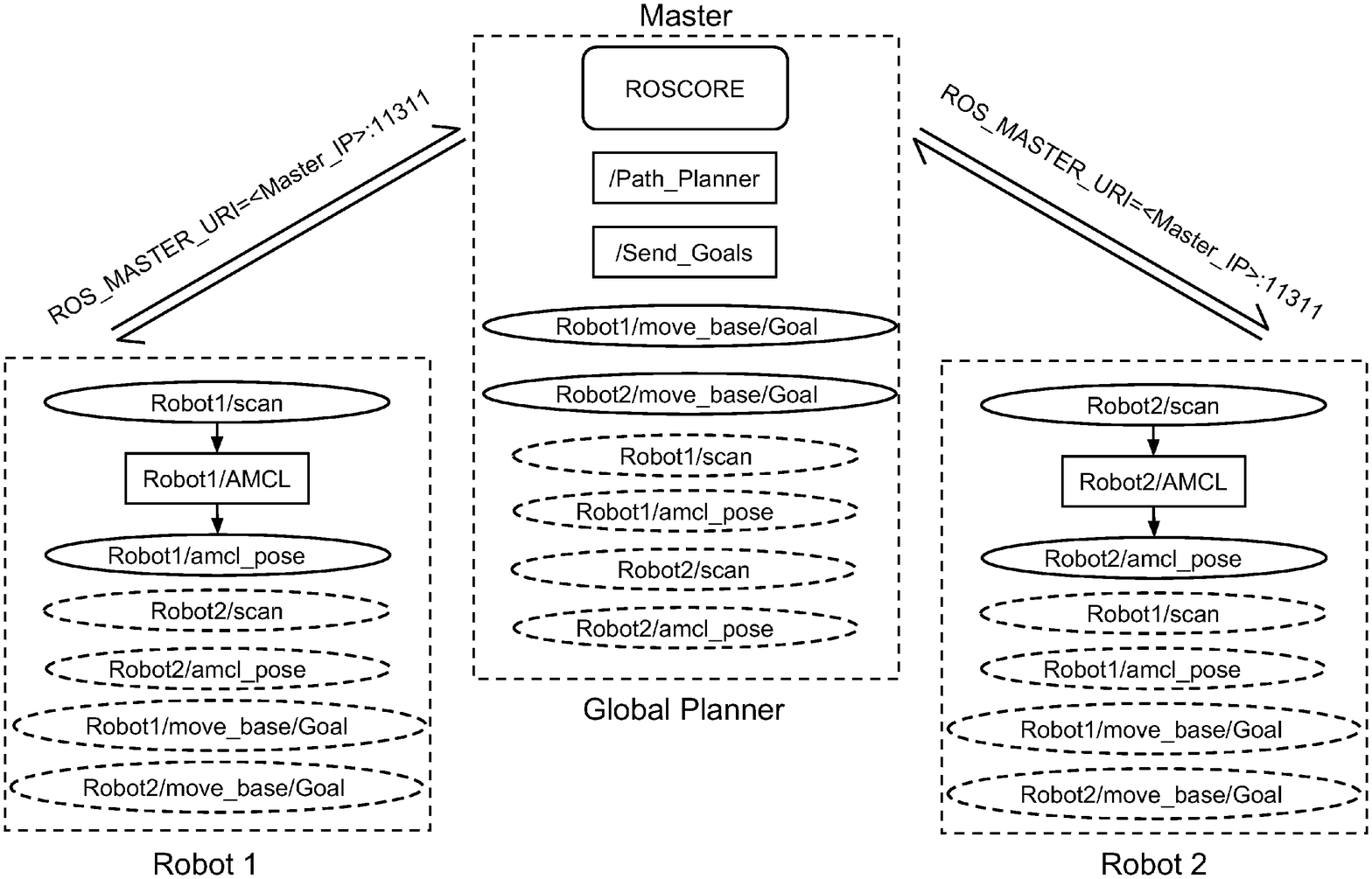}
 \caption{The figure shows two robots connected to a third system
 which is the master running ROSCORE. The master also runs the global
 planner. Rectangular boxes show nodes, solid oval shows topics on the
 machine, and dashed oval shows topics available for subscription from
 other machines.}
\label{fig:sms} 
\end{figure}

In a single master system, \texttt{ROSCORE} runs on one machine
which is called the master. Other nodes work in a distributed fashion
on different machines. The nodes can run anywhere on the network
except the driver nodes, which runs on the system that is directly
connected to the hardware.  All the nodes need to connect to  the
master. They connect via \texttt{ROS\_MASTER\_URI} which can be set in
\texttt{.bashrc} file of the respective machines as shown below. All
the machines in the network have a bidirectional connection with each
other. Also, the host IP and the master IP will be same in case of the
master machine.

\begin{flushleft}
  \small
  \rule{\columnwidth}{0.5pt}
  \begin{verbatim}
export ROS_MASTER_URI=http://<master_ip>:11311 
export ROS_HOSTNAME=<host_ip>
  \end{verbatim}
  \vspace{-0.5cm}
  \rule{\columnwidth}{0.5pt}
\end{flushleft}

Some of the common tasks like localization, mapping etc. runs on every
robot resulting in nodes with same name under \texttt{ROSCORE}. A
single launch file cannot be used to launch the nodes as it will
create a conflict and the previous running node will be overridden
with the new instance of the same name. This problem is resolved by
introducing \emph{namespace} and \texttt{tf\_prefix} tags in the
launch file as shown below.

\begin{flushleft}
  \small
 \rule{\columnwidth}{0.5pt}
 \begin{verbatim}
<launch>
<group ns="Robot1">
<param name="tf_prefix" value="Robot1" />
   .
   .
  <node pkg="<package_name>" 
  type="<node_type>" 
  name="<node_name>"> 
  <param name="<xyz>" 
  type="double" 
  value="<value_to_be_passed>" /> 
  </node>
   .
   .
</group>
</launch>
\end{verbatim}
\rule{\columnwidth}{0.5pt}
\end{flushleft}

The single master system can be set up by following the steps given below:
\begin{itemize}
  \item Setup \texttt{.bashrc} in each robot as shown above.
  \item Append suitable namespace and \texttt{tf\_prefix} to the nodes corresponding to each robot.
  \item Run \texttt{roscore} on the master.
  \item Launch each individual robot.
\end{itemize}

A single master system is handy for quick testing of
  algorithms on a single robot because of its simple setup process.
  Its simplicity, however, does not provide much advantage as the
  number of robots increase in the environment.  A schematic diagram
  of a working instance of single master system is shown in Figure
  \ref{fig:sms}. It shows one master running \texttt{roscore} and two
  client robots connected to the master over LAN. As one can see, all
  the topics from one robot is available for subscription by the all
  other robots as well. These topics are shown as dotted ellipse. The
  topics generated by the robot is shown as solid ellipses.  Making
  topics available to everyone all the time may lead to some security
  concern as one would like to have some control over who can access
  which topics. In other words, this would require additional overhead
  to restrict access to the topics of a given robot by the other.
  Secondly, the bandwidth requirement for a single master system with
  multiple robots is comparatively higher as all the topics are
  available over the network for subscription. Moreover, having a
  single master makes the whole system vulnerable because if
  \texttt{roscore} dies, service based communication between the nodes
  get stopped. Topic based communication can still work because once a
  connection between nodes is established via topics, \texttt{roscore}
  is no longer needed, but new topics cannot be created without
  \texttt{roscore} running.  Also, as the number of robots increase,
it becomes increasingly cumbersome to deal with conflict among similar
topics and namespace resolution.

\subsection{Multi Master System} \label{sec:mms}

\begin{figure}[!t] 
  \centering
  \includegraphics[width=\columnwidth]{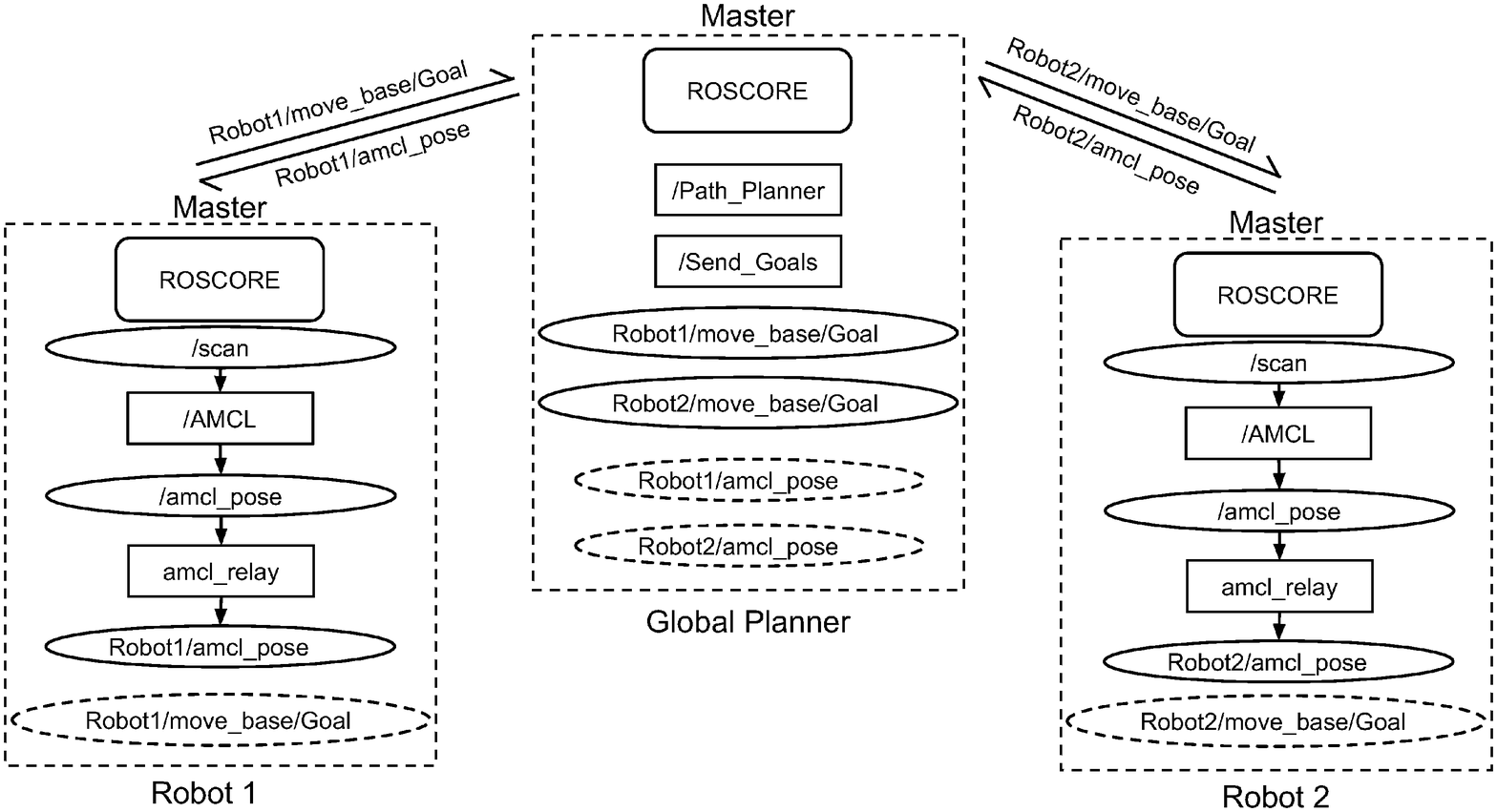}
\caption{A schematic view of a multi master system. The figure shows
  multiple \texttt{roscore}s running on different machines. In this
  configuration, there is no conflict among the topics with similar
  names as their visibility is limited to the machine running its own
  \texttt{roscore}. }
\label{fig:mms} 
\end{figure}

Many of the limitations of a single master system can
  be overcome by having multiple masters running their own independent
  \texttt{roscore} as shown in Figure \ref{fig:mms}. This makes the
  system robust as the failure of one will not lead to the failure of
  the complete system. Since the visibility of topics is limited to
  the scope of each \texttt{roscore} environment, there are no
namespace conflict with topics in a multi-master system. All the nodes
and services are local to that robot. However, it is possible to share
a minimum number of topics with other robots through remapping as and
when required. Since only a limited number of topics are shared, the
bandwidth required in a multi-master system is less compared to that
in a single master system for the same task.

To implement a multi-master System, a package called
\texttt{multimaster\_fkie} is needed \cite{juan2015multi} and can be
easily installed as shown below. This allows two important processes,
\texttt{master\_discovery} and \texttt{master\_sync} to run
simultaneously. The function of \texttt{master\_discovery} is  to send
multicast messages to the network so that all  \texttt{roscore}
environments become aware of each other. It also monitors the changes
in the network and intimates all ROS masters about these changes. The
other process called \texttt{master\_sync} enables us to select which
topics can shared between different \texttt{roscore}. Without
\texttt{master\_sync} node, no information can be accessed by other
\texttt{roscore}s. The following commands are required to be executed
to install and activate multi-master mode in each machine:

\begin{flushleft}
  \scriptsize
  \rule{\columnwidth}{0.5pt}
  \begin{verbatim}
$ sudo apt-get install ros-indigo-multimaster-fkie 
$ sudo sh -c "echo 0 >/proc/sys/net/ipv4/
                        icmp_echo_ignore_broadcasts" 
$ export ROS_MASTER_URI= http://<host_ip>:11311 
$ export ROS_HOSTNAME=<host_ip> 
$ roscore
$ rosrun master_discovery_fkie master\
                        _discovery_mcast_group:=224.0.0.1 
$ rosrun master_sync_fkie master_sync\
                        _sync_topics:=['topic_name']  
\end{verbatim}
\rule{\columnwidth}{0.5pt}
\end{flushleft}

It is to be noted that the host and master IPs are same on each
machine. This is unlike the single-master case where these two IPs
could be different for a given machine. The namespace conflict in
multi-master system can be avoided using a \emph{relay} node. The use
of relay node can be understood in the context shown in Figure
\ref{fig:mms}. The global planner needs to access pose data from Robot
1 and 2 for carrying out path planning. Each of these two robots
publish pose data to a topic called \texttt{/amcl\_pose} under their
respective \texttt{roscore}s. To avoid conflict, one has to relay the
\texttt{/amcl\_pose} of Robot 1 to the topic
\texttt{/Robot1/amcl\_pose} and that of Robot 2 to
\texttt{/Robot2/amcl\_pose} respectively. This can be done by
executing the following command on each of the robots:

\begin{flushleft}
  \scriptsize \rule{\columnwidth}{0.5pt}
\begin{verbatim}
$ rosrun topic_tools relay /amcl_pose /Robot1/amcl_pose
\end{verbatim}
\rule{\columnwidth}{0.5pt}
\end{flushleft}

As shown in the above figure, the global planner can now access 
these new topics called \texttt{/Robot1/amcl\_pose} and
\texttt{/Robot2/amcl\_pose} for obtaining their respective pose data.

Even though multi-master system saves us from several problems
encountered in a single master system, it still does not provide
solution to some other problems such as scalability, load balancing
and lower computation power. As number of robots increase, one needs
to reconfigure system files manually for each robot to enable
multi-casting. It does not make efficient use of the processing power
available because, by default, the processes are not distributed such
that load on each machine is balanced. Bandwidth usage in multi-master
system is still high compared to a cloud-based system due to the
difference in network protocols \cite{mohanarajah2015cloud} used by
different machines. In a multi-master system, each machine has a
limited on-board computational hardware which can not be augmented to
accommodate for higher demand in the run time. This limits the
usability of multi-machine system.

\subsection{Cloud Robotics System} \label{sec:crs} 

\begin{figure}[!t] 
  \centering
  \includegraphics[scale=0.3]{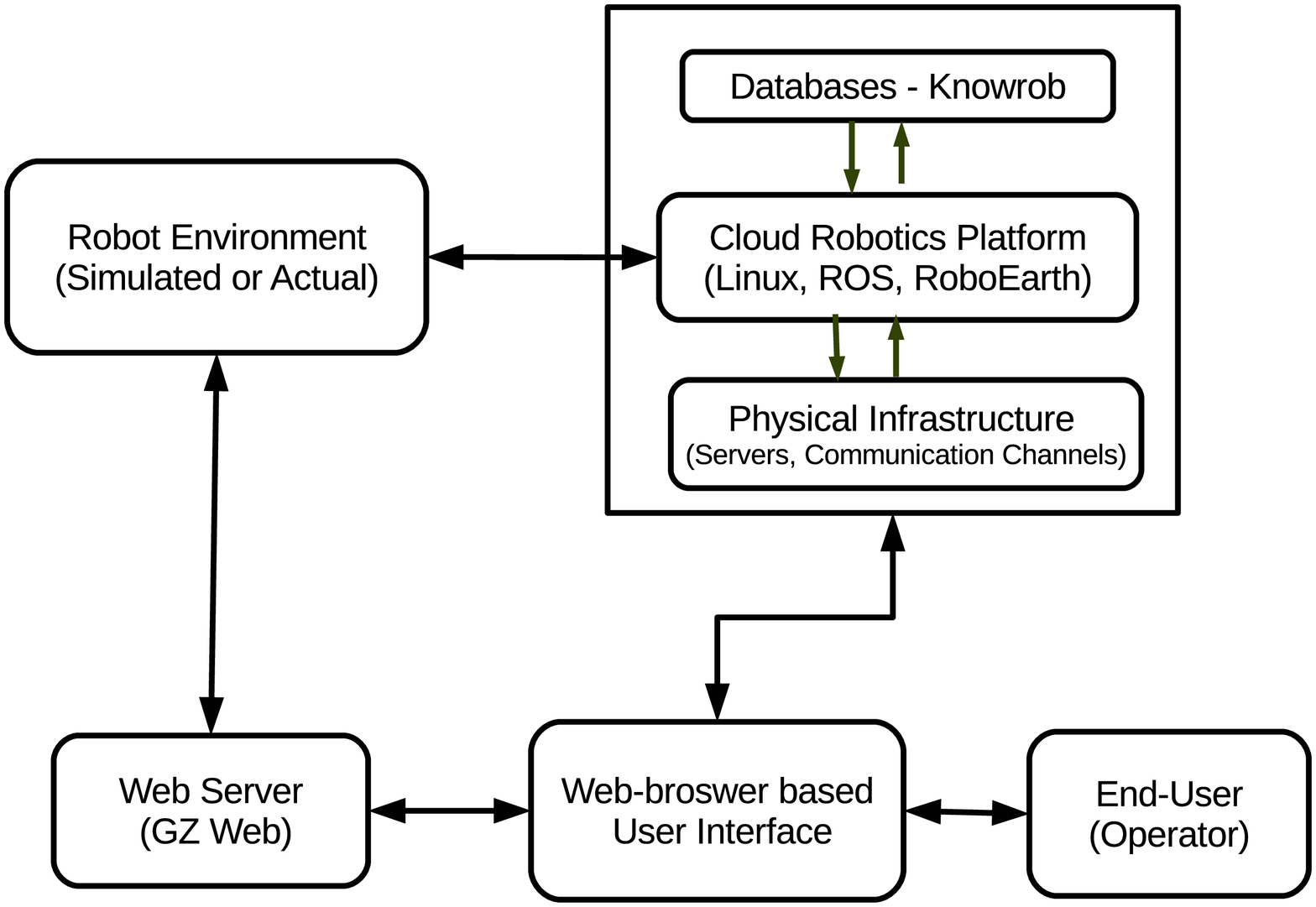}
  \caption{Block-diagram of implementation using a cloud robotics
  platform} 
  \label{fig:crsim_gzweb} 
\end{figure}

Many of the limitations of a multi-master system can
  be solved by having a cloud infrastructure to which the robots can
  offload computationally heavy tasks. In this paper, Rapyuta cloud
  robotics engine \cite{mohanarajah2015rapyuta}
  \cite{hunziker2013rapyuta} is used for implementing the fleet
  management system. As discussed earlier, it is a Platform as a
  Service (PaaS) framework suitable for developing robotic
  applications. The schematic of such an implementation is shown in
  Figure \ref{fig:crsim_gzweb}. It shows four main components: (1) a
  cloud server which includes both software as well as hardware
infrastructure; (2) Physical or simulated Robots and their working
environment. (3) an user interface for interacting with the system and
(4) an operator or an ERP system to provide goals for the system.

\begin{figure}[!t]
  \centering 
  \includegraphics[width=\columnwidth]{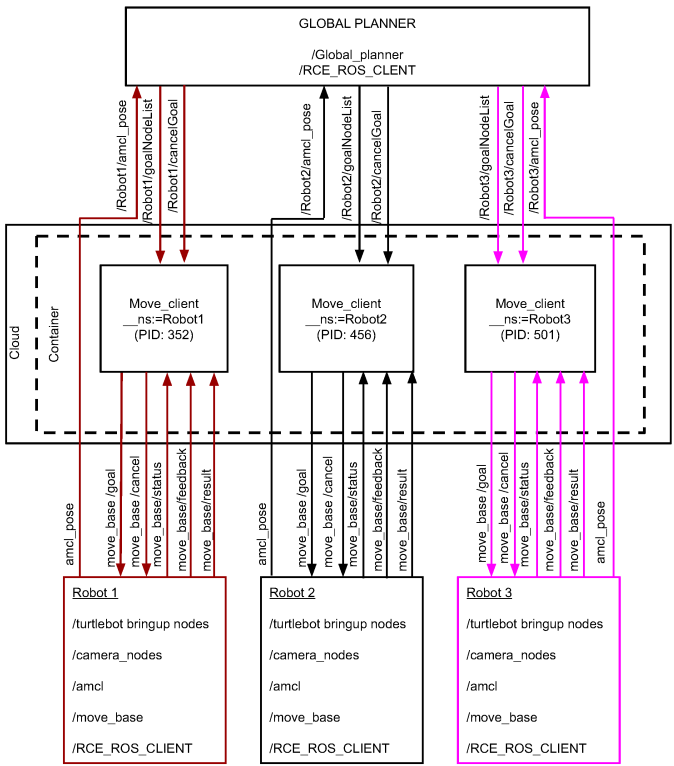}
\caption{The process nodes and topics required for implementing the
  fleet management system using Rapyuta cloud robotics engine. The
system shows four agents (three robots and one global planner)
interacting with each other through a cloud server. In this
implementation, only a single container is used to execute all
relevant processes. The arrow heads show the direction of information
flow through topics between different nodes. } 
\label{fig:ros_graph} 
\end{figure}

The inner working of this cloud-based implemented
  could be better understood by studying the Figure
  \ref{fig:ros_graph} that provides a process level overview of the
  system showing nodes, topics and interconnection pathways among
  various modules of the fleet management system. The figure shows a
  five agent system implemented using four physical machines (three
  robots and a server).  Each robot runs processes for localization
  and autonomous navigation through nodes \texttt{/amcl}
  and \texttt{/move\_base} respectively. The processes related to Rapyuta cloud
  robotics engine runs on the server machine. It also runs processes for global
  planner which generates paths for the robots. In a general scenario,
  the global planner and all related optimization algorithms can run
  on a separate physically machine on the network. Hence, it is shown
  as a separate block in the Figure \ref{fig:ros_graph} similar to the
blocks corresponding to robots.

As shown in this figure, the global planner publishes
  data into two types of topics. The first topic is
  \texttt{/goalNodesList} which provides paths generated by the
  planner in the form of an array of grid block numbers. Each robot
  subscribes to its corresponding \texttt{goalNodeList} to know the
  cell locations that it needs to traverse. The second topic, called
  \texttt{/cancelGoal}, is a binary number which indicates whether the
  current goal locations received from the global planner is to be
  discarded by the robot or not. The binary value for the
  topic \texttt{/cancelGoal} for a given robot is set if a cell on its
  path is blocked either by an user or by an obstacle detected by the
  robot sensors. The grid cells could also be blocked by an ERP
  (Enterprise Resource Planning) system indicating non-traversable
  regions in the environment.  Whenever the value for
  \texttt{/cancelGoal} is set, the robot discards previously
  received goal locations and uses new values available at the
  corresponding \texttt{/goalNodesList} topic.  These topics are
  subscribed by the respective \texttt{move\_client} nodes on the
  cloud which, in turn, publish necessary topics for use subscription
by the physical robots.

Before going further, a brief understanding of Rapyuta organization
will be useful for understanding the configuration steps described
later. Rapyuta has the following four main components
\cite{mohanarajah2015rapyuta}. (1) \emph{Computing environments} are
the Linux containers \cite{dua2014virtualization}
\cite{joy2015performance} used for running various ROS based robot
applications; (2) \emph{Communication protocols:} are the standard
protocols used for internal and external communication between cloud,
container and robot processes. (3) \emph{Core Task Set:} for managing
all process and tasks. They are further divided into three groups,
namely, robot task set, environment task set and container task set.
(4) \emph{Command Data Structures:} are the necessary formats used for
various system administration activities.

The setup process for the cloud robotics based fleet management system
involves two main step: 
\begin{itemize}
  \item Create configuration files providing details of interaction between cloud and robots. 
  \item Launch these files using system commands on server as well as robot clients. 
\end{itemize}
In the remaining part of this section, we provide the details of
configuration on server as well as the clients.

\subsubsection{Configuration of Cloud Server} \label{sec:cloudcfg}

\begin{figure*}[!t] 
  \centering
  \includegraphics[width=\textwidth]{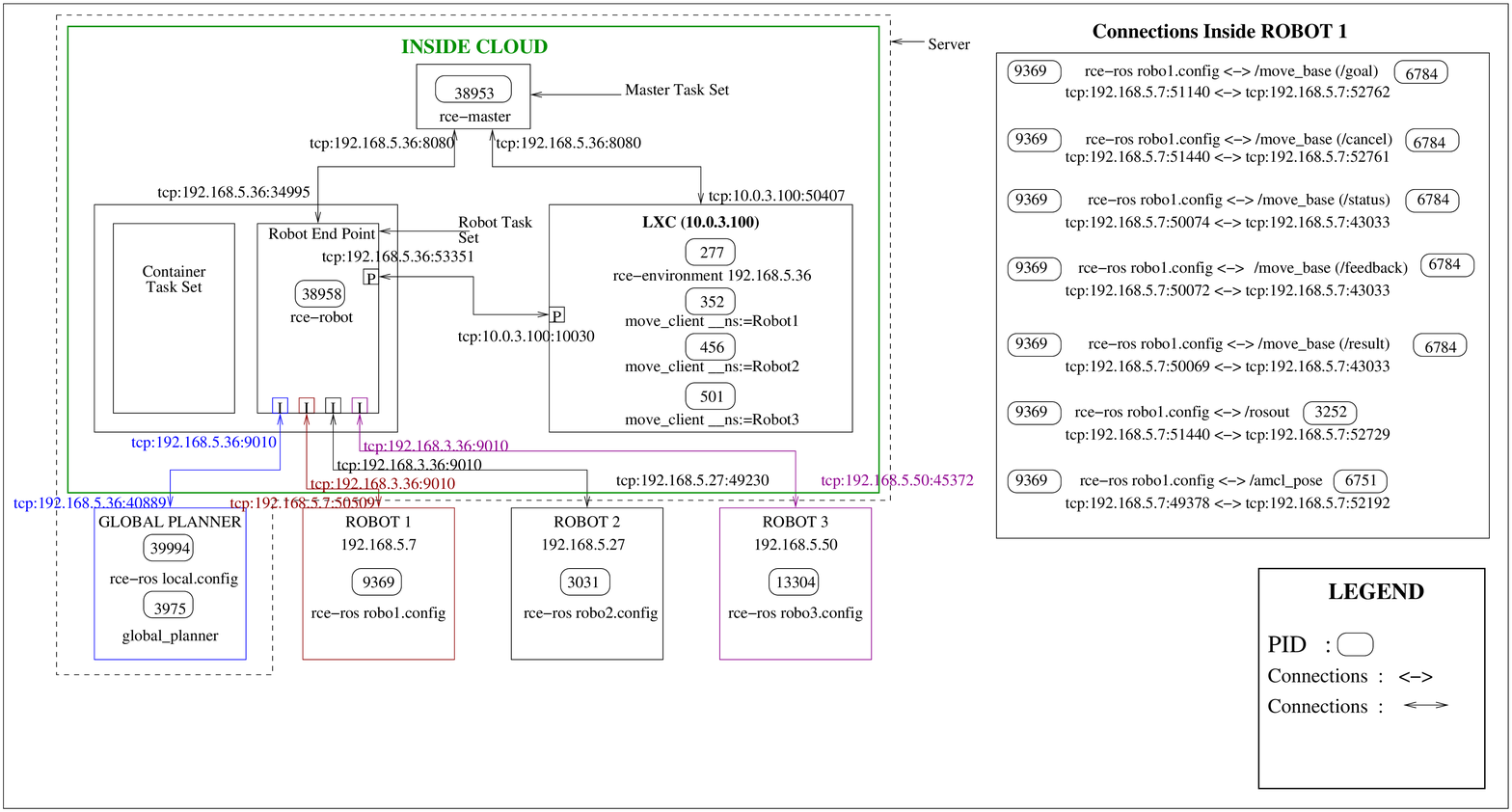}
\caption{Configuration for multiple AMRs in a Rapyuta-based fleet
  management system. The dashed line shows the server system where
  Rapyuta cloud engine is running along with a ROBOT process called
  global planner. Three AMRs are represented by the three blocks
  termed as Robot 1 , Robot 2 and Robot 3. On right hand side
  processes inside ROBOT 1 are shown and their interaction with
  rce-ros process to send data. As shown in figure, move\_base process
  having PID 6784 communicates with rce-ros process having PID 9369
  through system assigned ports.}
\label{fig:cloud_config} 
\end{figure*}

The configuration for the cloud-based fleet management system is shown
in Figure \ref{fig:cloud_config}. The dotted box shows the activities
within the cloud server.  The first process which needs to be started
on the server is the \emph{Master Task Set} which controls and manages all
other processes on the cloud. It takes up an IP called
\texttt{master\_ip} and listens on port 8080. This process is started
by executing following Linux command:

\begin{flushleft}
  \small
  \rule{\columnwidth}{0.5pt}
  \begin{verbatim}
$ rce-master
\end{verbatim}
\rule{\columnwidth}{0.5pt}
\end{flushleft}

The next process which needs to be started on the
server is the \emph{Robot Task Set} which is responsible for managing
communication  with physical robots. It can be started using the
following command:  
\begin{flushleft}
  \small
  \rule{\columnwidth}{0.5pt}
  \begin{verbatim}
$ rce-robot <master_ip> 
\end{verbatim}
\rule{\columnwidth}{0.5pt}
\end{flushleft}

The third task which needs to
be started is the \emph{Container Task Set} responsible for managing
containers which are the basic computing environment on the cloud. The
corresponding command is:

\begin{flushleft}
  \small
  \rule{\columnwidth}{0.5pt}
  \begin{verbatim}
$ sudo rce-container <master_ip> 
\end{verbatim}
\rule{\columnwidth}{0.5pt}
\end{flushleft}

Each Linux container (LXC) takes up its own IP and port to communicate
with master. Linux containers need not be collocated with the Rapyuta
server (\texttt{rce-master}) and can run on any other machine on the
network. It is also possible to have multiple containers. The linux
containers are capable of running standard available ROS nodes or
user-created nodes to perform a specific task.  Inside each Linux
container, lies the fourth and final core task set known as
\emph{Environment Task Set}. This task set allows the ROS nodes
running within the container to communicate with other nodes running
on other Linux containers and robots on the network. The configuration
for these environment tasks for containers are provided in the
configuration files used by the individual clients as will be
explained in the next section.

The Figure \ref{fig:cloud_config} also shows two main types of
connection for communication among various processes. One is for
internal communication within different Rapyuta processes and, the
other one is for external communication between Rapyuta processes and
robots. Internally, Rapyuta communicates over UNIX Sockets. For
instance, the master task set uses port \texttt{8080} for
communication and is referred to as an \texttt{internal\_port}.
The processes within the Linux container communicate with robot end
points through communication ports or \texttt{comm\_port}. The
corresponding port number is \texttt{10030} and is represented by the
letter `P' (stands for ports) in the above figure. The robot endpoints
provide \emph{interfaces} for converting external format (e.g. JASON)
into internal format of robots (e.g. ROS messages). On the other hand,
\emph{ports} are used for internal communication between endpoint
processes.  The external communication between Rapyuta processes and
robots uses web-socket protocol. This communication is over
\texttt{9010} port which is also knows as \texttt{ws\_port} or
\texttt{websocket\_port}.  Readers can refer to
\cite{mohanarajah2015rapyuta} for more details. The figure also shows
the process IDs (PID) for all related topics and nodes.

\subsubsection{Configuration for Robots} \label{sec:robcfg}

In order to demonstrate the working of the system, Turtlebots
\cite{turtlebot} are used as autonomous mobile robot (AMR) platforms
for our fleet management system. After setting up the cloud, robot
processes are required to be started on each robot. Each robot is made
alive within ROS environment by using \texttt{turtlebot\_bringup}
command. Other functionalities of the robot (autonomous navigation,
obstacle avoidance, localization etc.) are activated through a
standard ROS launch file. The connection between robot and Rapyuta is
established using \texttt{rce-ros} command with a local configuration
file available on each robot. The basic commands for setting up robots
are as follows:

\begin{flushleft}
  \small
  \rule{\columnwidth}{0.5pt}
  \begin{verbatim}
$ turtlebot_bringup
$ sudo rce-ros robot1.config
$ roslaunch botmotion.launch
\end{verbatim}
\rule{\columnwidth}{0.5pt}
\end{flushleft}

The configuration files are written in JSON and are used for sending
request instructions to master task set for establishing connection
with the cloud. The configuration file for each robot has the
following four main components: (1) Containers, (2) Nodes, (3)
Interfaces and (4) Connections. Other than this, the first part of the
configuration file is used to send HTTP request to the cloud. This part
appears as shown below:

\begin{flushleft}
  \small
  \rule{\columnwidth}{0.5pt}
  \begin{verbatim}
"url":"http://192.168.5.36:9000/", 
"userID" :"testUser",  
"password" : "testUser", 
"robotID"  : "testRobot_1",   
\end{verbatim}
\rule{\columnwidth}{0.5pt}
\end{flushleft}

As shown above, the request is sent on port \texttt{9000} and in
response, Rapyuta sends the endpoint's URL to the robot as a JSON
encoded response. This received URL is used by the robot to connect
with the cloud through port \texttt{9010}. These ports are configured
at the time of installation.   Upon establishing the connection the
robot requests for container creation and it is done by the following
block in the configuration file:

\begin{flushleft}
  \small
  \rule{\columnwidth}{0.5pt}
  \begin{verbatim}
"containers": ["cTag" : "cTag_01" ] 
\end{verbatim}
\rule{\columnwidth}{0.5pt}
\end{flushleft}

This creates a container inside Rapyuta having a unique tag
provided by the key \texttt{"cTag"}. Each container starts with the
necessary processes or daemons like \texttt{roscore}, \texttt{sshd},
etc. and looks for the nodes which needs to be run inside the
container. This information is provided in the `node' block in the
configuration file as shown below: 

\begin{flushleft}
  \small
  \rule{\columnwidth}{0.5pt}
  \begin{verbatim}
  "nodes": [ 
  "cTag" : "cTag_01",
  "nTag" : "move_client_node_1", 
  "pkg"  : "move_client",
  "exe"  : "move_client_pthread", 
  "args" : "/Robot1/goalNodesList/Robot1, 
                     /cancelGoal, Robot1/map", 
  "namespace" : "Robot1" 
  ... ]
\end{verbatim}
\rule{\columnwidth}{0.5pt}
\end{flushleft}

The key \texttt{"cTag"} refers to the name of the container where
these nodes are to be created, \texttt{"nTag"} specifies the name for
the node, \texttt{"pkg"} tells the master task set about the needed
packages. The key \texttt{"exe"} tells the name of the executable,
\texttt{"args"} contains the arguments to be passed and
\texttt{"name-space"} segregates the processes inside the container
giving us the flexibility to run multiple copies of the same executable
independently inside a container. 

Once the nodes are up, it is necessary to define interfaces for each
robot. Interfaces primarily refer to various kinds of sensor data that
will be shared with the cloud or other robots in the network. This is
specified by the following block in the configuration file: 

\begin{flushleft}
  \small
  \rule{\columnwidth}{0.5pt}
  \begin{verbatim}
  "interfaces": [ { 
  "eTag" : "cTag_01", 
  "iTag" : "amclPoseReceiver_1", 
  "iType" : "PublisherInterface", 
  "iCls" :  "geometry_msgs/
               PoseWithCovarianceStamped",
  "addr" : "/Robot1/amcl_pose" } 
\end{verbatim}
\rule{\columnwidth}{0.5pt}
\end{flushleft}

The key \texttt{"eTag"} refers to the endpoint tag which is either a
robot end or a container end and accordingly, a robot ID or a
container tag can be mentioned as its value. The key \texttt{"iTag"}
is the interface tag and is unique in the scope of an endpoint tag.
\texttt{"iType"} defines the type of the interface tag which can be
subscriber, publisher, service client or service provider as defined
by Rapyuta \cite{rapyuta_dev_res}.  \texttt{"iCls"} refers to the
class name and it defines the message type for publisher or subscriber
and \texttt{"addr"} is the address of ROS topic. After defining the
interfaces, it is necessary to specify the connections between various
endpoints as shown in the following block:

\begin{flushleft}
  \small
  \rule{\columnwidth}{0.5pt}
  \begin{verbatim}
  "connections" : [ { 
  "tagA" : "cTag_01/amclPoseReceiver_1", 
  "tagB" : "testRobot_1/amclPoseSender_1" }, 
\end{verbatim}
\rule{\columnwidth}{0.5pt}
\end{flushleft}

This part establishes the connection between interfaces. The points to
be connected are defined as \texttt{"tagA"} and \texttt{"tagB"}.


\section{Simulations \& Experiments} \label{sec:sim}

In this section we will provide details of how
  different components of fleet management system work. The modules
  that are being discussed here include global planner, gazebo
  simulation model and the web-based user interface. 

  \subsection{Global Planner} \label{sec:gplan}

\begin{figure}
  \centering
  \begin{tabular}{cc}
    \subfloat[]{\label{f:p1}\includegraphics[width=1.5in]{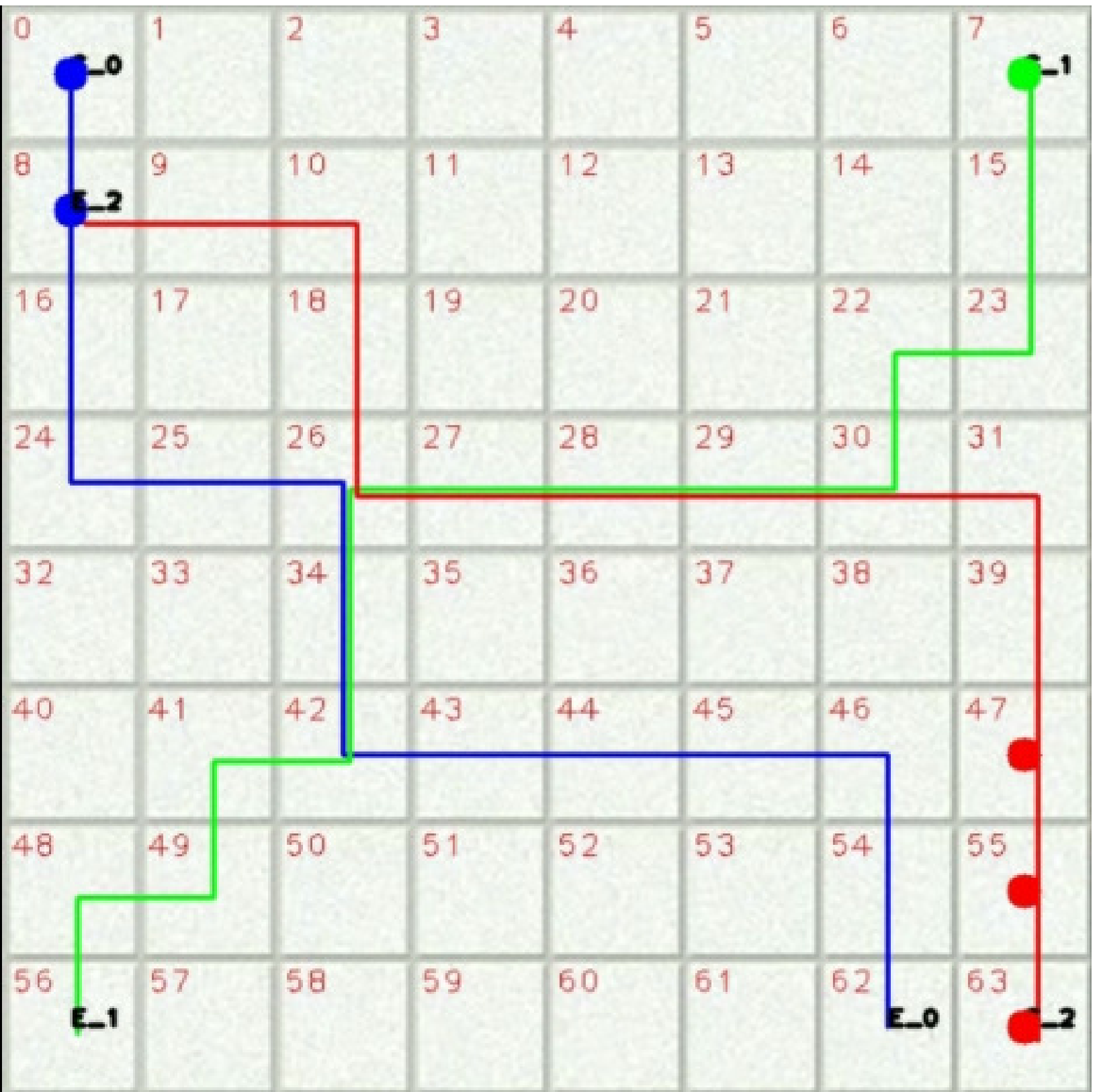}} & 
    \subfloat[]{\label{f:p2}\includegraphics[width=1.5in]{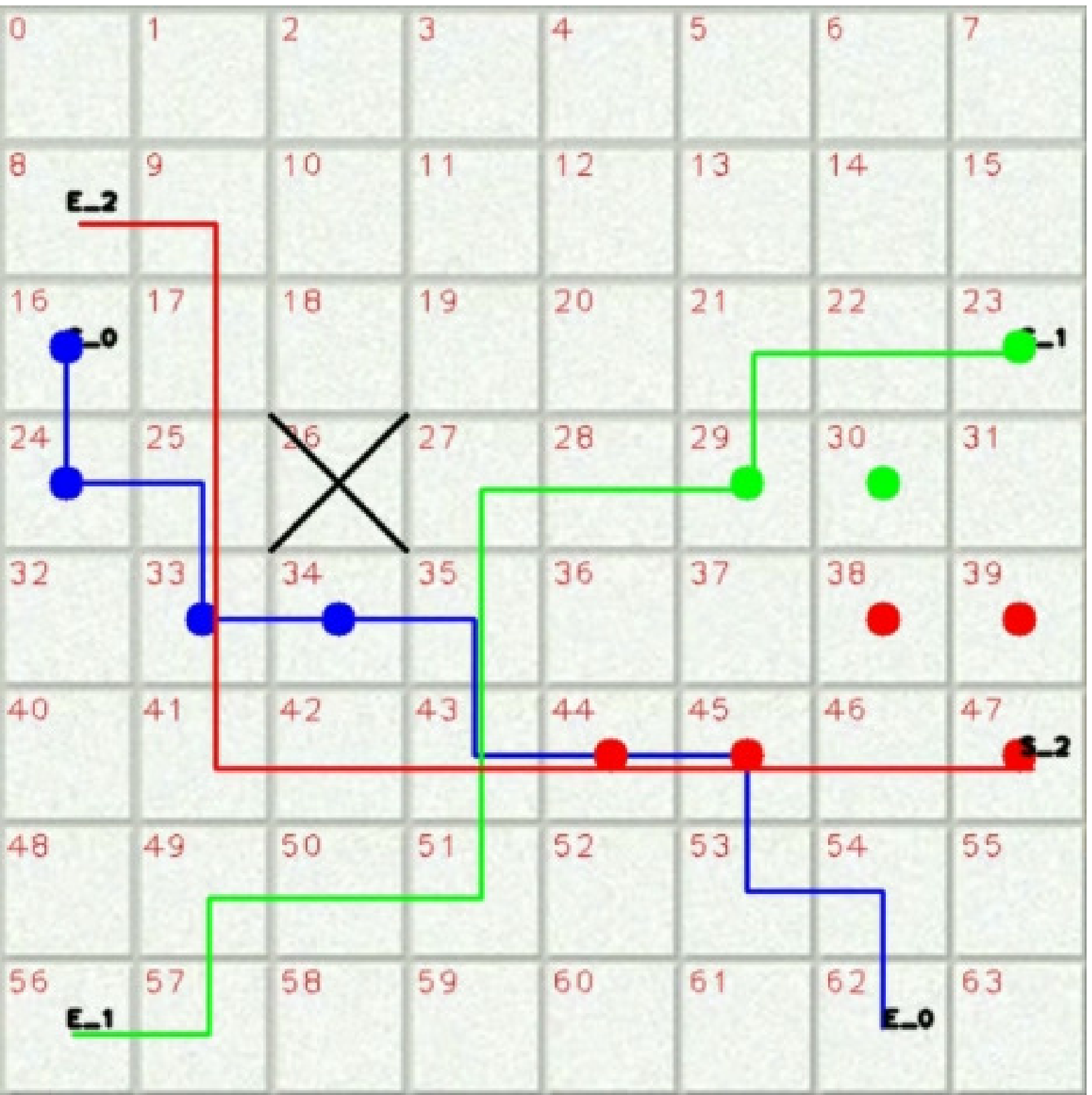}}  
  \end{tabular}
  \caption{Paths generated by Global Planner: (a) Paths for three robots
    obtained without any obstacles. Circular dots show the location of the
    robot as it traverses this path. (b) Shows new paths generated by the
    global planner once the user blocks the cell number  26. Grid cells
  can also be blocked when a robot detects an obstacle.}
  \label{fig:planner}
\end{figure}

  As discussed earlier, the global planner is
    responsible for generating paths for robots between their
    current locations and the target destinations provided by the
    operator. It receives the location information from each of the
    robots, the destination information for these robots from the
    operator and, uses the latest map to generate necessary paths for
    the robots. In its simplified form, it implements a Dijkstra
    algorithm \cite{goldberg2005computing}\cite{goldberg2006reach}
    \cite{skiena1990dijkstra} on a grid map to find shortest path
    between two cells as shown in Figure \ref{fig:planner}. In this
    figure, the robots are represented by filled circles. The start
    and end destinations of these robots are represented by the symbol
    pair $\{S_i, E_i\},i=1,2,\dots,N$ where $N=3$ in this case. The Figure
    \ref{f:p1} shows the case when no obstacles are present in the
    map. As soon as the path information is transmitted to the robots,
    they start following their respective paths as shown by the trail
    of circular dots on their paths. The Figure \ref{f:p2} shows the
    case when an obstacle is created (or detected) in the cell number
    26 at any time during this motion. This results in
    generation of new paths by the global planner. In a simulated
    environment, the robots can react instantaneously to this change.
    However, the robots may take some in a real world scenario due to
    factors like communication delay and inertia of motion as shown in
    this figure. The global planner may also include several other
    factors such as, battery life of robots, additional on-board
    sensor or actuator on robots (in case of a heterogeneous scenario)
    and other environmental conditions to solve a multi-objective
    optimization problem to generate these paths. Our purpose in this
    paper has been to demonstrate the working of a complete fleet
    management system which invariably requires such a centralized
    planner for task allocation and towards this end, we pick up the
  simplest path planner as an example. Readers are free to explore
other planners in the same context.

%

\subsection{Simulation Environment}\label{sec:gsim}

The simulated environment for the fleet management system is created
using Gazebo \cite{koenig2004design} \cite{gazebo}, which is an
open-source software well integrated with ROS. The steps required for
this on a Ubuntu Linux environment are as follows: 

\begin{itemize}
  \item Create "\texttt{model.config}" and "\texttt{model.sdf}" file
    and place them in a folder preferably in folder \texttt{.gazebo/models/}.
  \item  Create a launch file similar to \texttt{empty\_world.launch}
    and set "\texttt{world\_name}" argument as the address of your
    newly created world by using the following command: 
\end{itemize}

    \begin{flushleft}
      \small
      \rule{\columnwidth}{0.5pt}
      \begin{verbatim}
$ roslaunch empty_world.launch world_name:=
            'address of newly created world'
      \end{verbatim}
      \vspace{-0.5cm}
      \rule{\columnwidth}{0.5pt}
    \end{flushleft}

\begin{figure}[!h] \centering
  \includegraphics[scale=0.3]{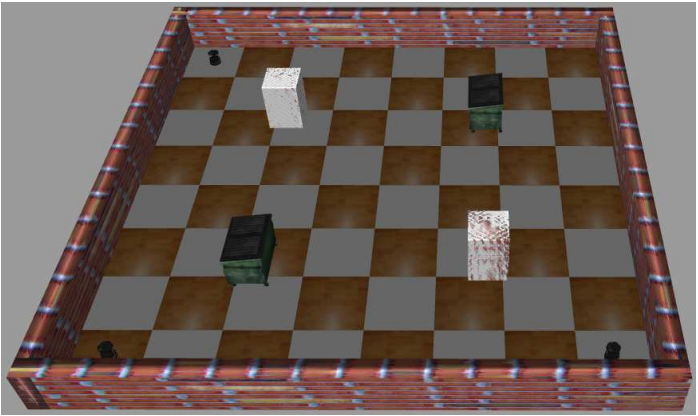}
  \caption{The simulated Gazebo environment for the fleet management
  system. There are four obstacles and three robots spawned in
environment.}
  \label{fig:gazebo_interface}
\end{figure}

The resulting simulated environment is shown in Figure
\ref{fig:gazebo_interface}. It also shows three obstacles (cuboidal
blocks) and three robots which are spawned in the environment. The
grid cells on the floor correspond to the grid map used by the global
planner shown in Figure \ref{fig:planner}. Whenever an user blocks a
cell in the grid map, a cuboidal block is spawned in the Gazebo
environment. The slow performance of Gmapping algorithm in Gazebo
simulation might be overcome by tweaking some scan matching parameters
as shown below:
    \begin{flushleft}
      \small
      \rule{\columnwidth}{0.5pt}
      \begin{verbatim}
<param name="minimumScore" value="10000"/> 
<param name="srr" value="0"/> 
<param name="srt" value="0"/> 
<param name="str" value="0"/> 
<param name="stt" value="0"/> 
<param name="particles" value="1"/> 
      \end{verbatim}
      \rule{\columnwidth}{0.5pt}
    \end{flushleft}

\subsection{Web User Interface}\label{sec:webui}

\begin{figure}[!h]
  \centering
  \begin{tabular}{c}
    \subfloat[Browser interaction with Gazebo and ROS]{\label{fig:gzweb}\includegraphics[scale=0.45]{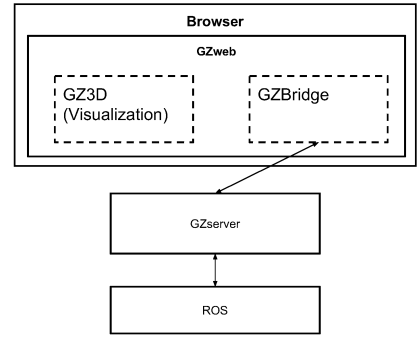}} \\ 
    \subfloat[Web interface for cloud based fleet management system]{\label{fig:webui}\includegraphics[scale=0.2]{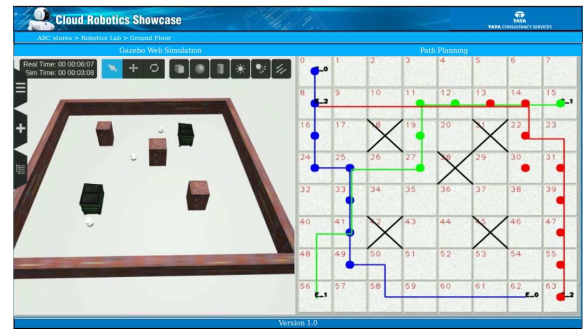}}  
  \end{tabular}
  \caption{Simulated Environment for the multi-robot fleet management system. } 
  \label{fig:web}
\end{figure}

A web interface is also built to interact with the
  robots and view the robot movement. This is shown in Figure
  \ref{fig:webui}. The figure shows two windows - one for visualizing
  the robot motion in a simulated environment and an interactive grid
  map for user interaction. The user can block out cells to spawn
  obstacles in the simulated environment and select starting position
  for robots. The interface is created using
  \texttt{Gzweb}\footnote{\burl{http://gazebosim.org/gzweb}} which is
  a web graphics library (WebGL) for Gazebo simulator. Like
  \texttt{Gzclient}, it is a front-end interface to \texttt{GZserver}
  and provides visualization of the simulation. It is a comparatively
  thin and light weight client that can be accessed through a web
  browser. The organization of this interface is shown in Figure
  \ref{fig:gzweb}.  \texttt{Gzweb} uses \texttt{Gz3D} for
  visualization and interacts with \texttt{Gzserver} through
  \texttt{Gzbridge}. \texttt{Gzserver} which forms the core of the
Gazebo simulator can interact with user programmes written with ROS
APIs. This web-based interface makes the whole system
platform-independent where an user can access the system over internet
without having to worry about installing the pre-requisite software on
his/her system. 


\subsection{The Experimental Setup} \label{sec:expt}

\begin{figure}[!h] 
  \centering
  \begin{tabular}{c}
    \subfloat[Actual robot setup in a
    laboratory]{\label{fig:real_robots}\includegraphics[width=0.4\textwidth,
    height=3cm]{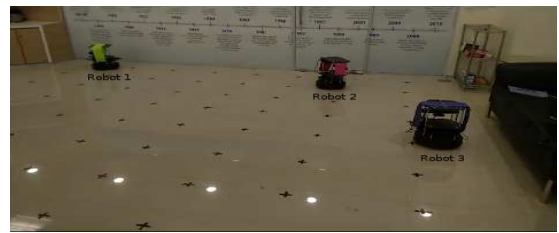}} \\ 
    \subfloat[RVIZ visualization of map and
    robots]{\label{fig:real_slam}\includegraphics[width=0.4\textwidth,
    height=3cm]{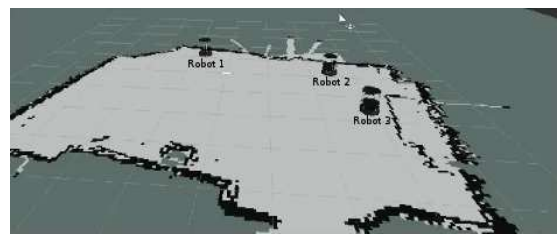}}  
  \end{tabular}
  \label{fig:expt_setup}
\caption{Experimental setup for testing the working of the fleet
management system.} 
\end{figure}

In this section, we provide details of our real world
  experiment with physical robots. Three Turtlebots \cite{turtlebot}
  are used as autonomous mobile robots (AMR) in a lab environment as
  shown in Figure \ref{fig:real_robots}. The map of the environment is
  created by using Gmapping SLAM algorithm available with ROS
  \cite{santos2013evaluation}. The map generated is shown in Figure
  \ref{fig:real_slam}.  Each of the robots run AMCL \cite{amcl}
  \cite{zaman2011ros} to localize itself in the map. It also runs an
  obstacle avoidance algorithm that uses on-board Kinect depth range
  information to locate obstacles on the path. These programs are run
  on a low power Intel Atom processor based netbook with 2 GB RAM that
  comes with these robots. The map is divided into equispaced $8\times
  8$ grid to match with the grid up used by the global planner shown
  in Figure \ref{fig:planner}. The server is a 12 CPU machine with
  Intel Xeon processor with 48 GB of RAM and 2 TB of storage space.
  The robots and server communicate over a local wireless LAN. The
  complete video of the experiment \cite{demo} as well as the source
  codes \cite{democode} are made available online for the convenience
  of users.

\section{Performance Analysis} \label{sec:perf}

The performance of each of the three modes of operation is analyzed by
performing two different experiments.  The details of the experiment
and the resulting analysis is provided in this section. 

\subsection{Experiment 1} \label{sec:expt1}

\begin{figure}[!h]
\centering
\begin{tabular}{c}
  \subfloat[Single Master System (SMS)]{\label{f:ssm}\includegraphics[scale=0.3]{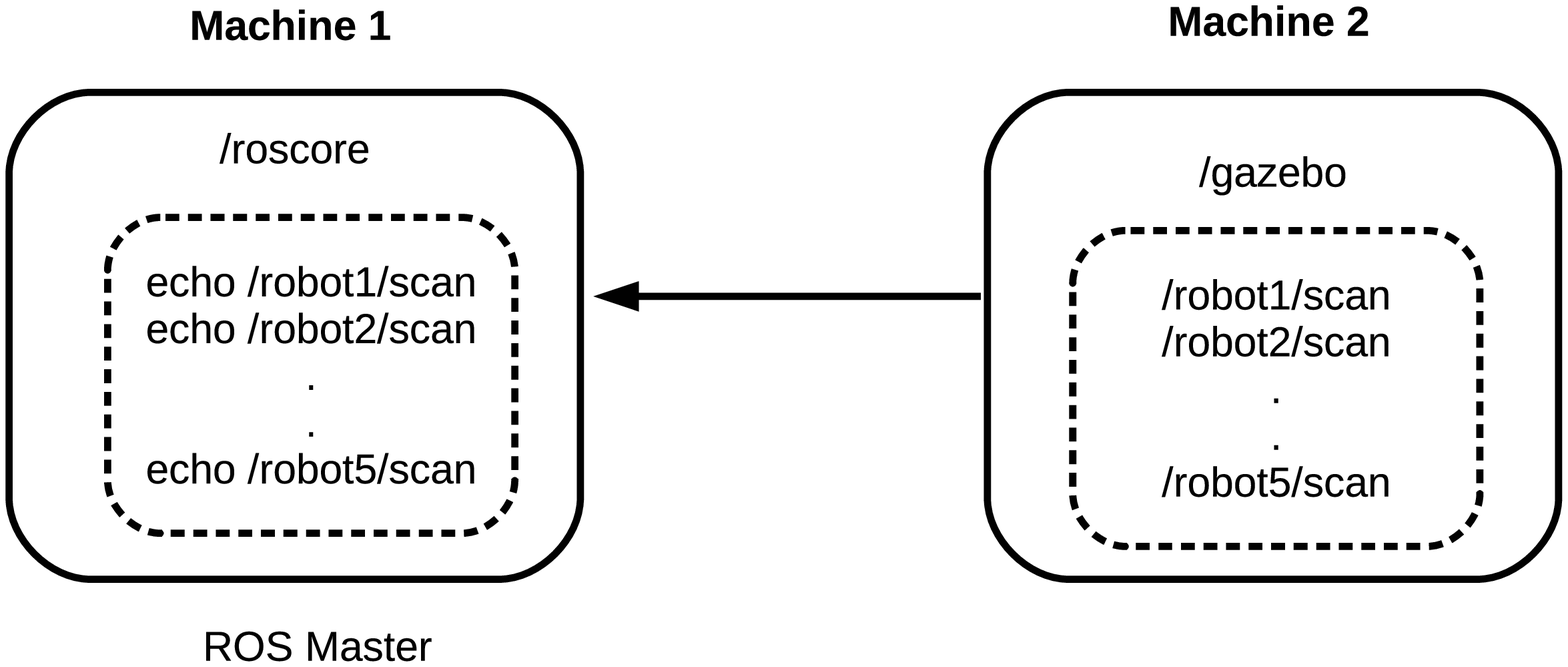}} \\
  \subfloat[Multi-master System (MMS)]{\label{f:smm}\includegraphics[scale=0.3]{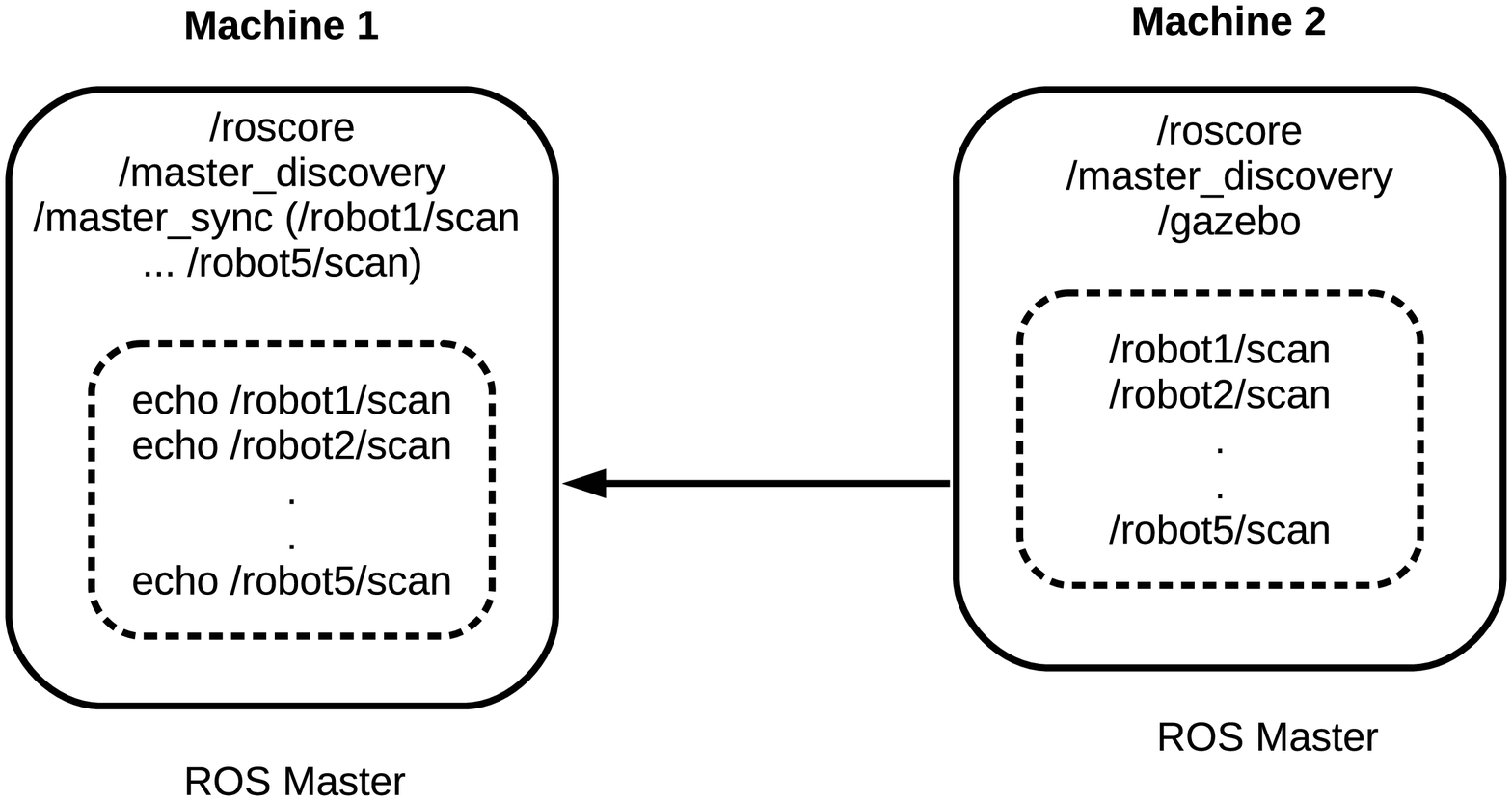}} \\ 
  \subfloat[Cloud Robotics System (CRS)]{\label{f:crs}\includegraphics[scale=0.3]{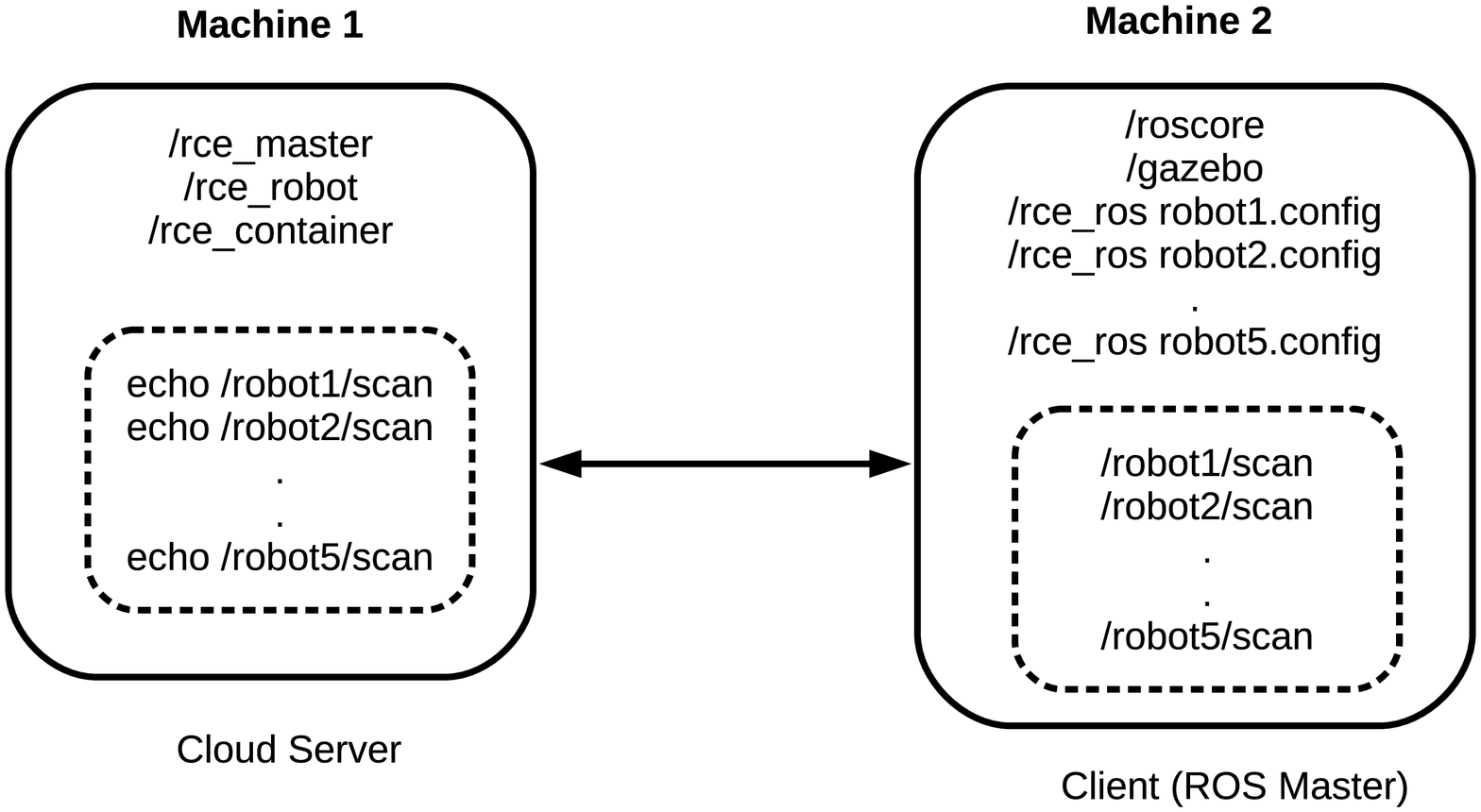}} 
\end{tabular}
\caption{Schematic of simulation experiment carried for analyzing the
performance of each of the three modes of implementation. The figure
shows the essential nodes running and topics available for
subscription on each of the machines.}
\label{fig:simexpt}
\end{figure}

The schematic of machine configuration used for this experiment is
shown in Figure \ref{fig:simexpt}. It shows two physical machines in
the network connected to each other through Wireless LAN.  The figure
\ref{fig:simexpt}(a) shows the single-master mode where \emph{Machine
1} acts as the master running \texttt{roscore}. Machine 2 runs Gazebo
simulation environment as explained in Section \ref{sec:gsim} and
spawns five Turtlebots in it.  Machine 1 apart from running
\texttt{roscore} subscribes to the Kinect scan data from these robots
and prints them on a terminal console. The Figure \ref{fig:simexpt}(b)
shows the multi-master mode of operation where both machines run their
own \texttt{roscore} processes. As before, the machine runs its own
Gazebo simulation environment and spawns a set of five Turtlebots.
Each of the machines run \texttt{master\_discovery} node to detect
other masters in the network. The machine 1 runs the
\texttt{master\_sync} node to subscribe to the scan data from all
robots running on the other machine. The Figure \ref{fig:simexpt}(c)
shows the cloud-based mode of operation where Machine 1 acts as the
cloud server running Rapyuta nodes such as \texttt{/rce\_master},
\texttt{/rce\_robot} and \texttt{/rce\_container}. Similar to the
previous case, the other machine runs its own Gazebo simulation
environment and spawns its own set of five Turtlebots. In addition
these machines also run \texttt{/rce\_ros} nodes for each of the robot
in order to establish connection with the cloud. In this case as well,
the server subscribes to the scan data from all robots from both the
machines through a container process.

\begin{figure}[!h] \centering
  \includegraphics[width=3.5in]{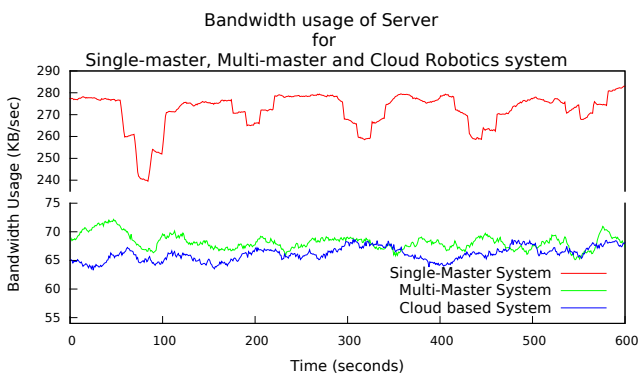}
  \caption{Network usage for Machine 1 for three modes of
  implementation.  The Machine 1 echoes robots scan data in each of
three cases. It shows that single master system generates more traffic
compared to other two configurations under identical conditions. }
    \label{fig:netuse} 
  \end{figure}

\begin{figure}[!h] 
  \centering
  \includegraphics[width=3.5in]{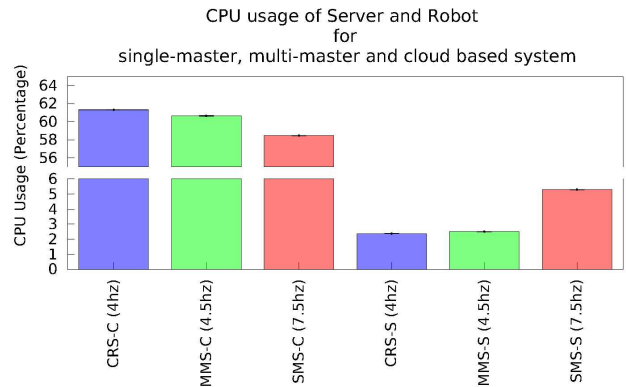}
\caption{CPU resource usage on server as well as client for three
  modes implementation: single master system (SMS), Multi-master
  system (MMS) and Cloud Robotics System (CRS). An additional letter
  `S' or `C' is used to represent a server or a client machine
respectively. It also shows the default publishing rate of messages on
each topic. }
  \label{fig:cpu} 
\end{figure}

The relative performance of each of the modes of implementation can be
analyzed by studying the two parameters, namely, network usage and CPU
usage of the machines as explained below. The network usage for
Machine 1 for all the three configurations is shown in Figure
\ref{fig:netuse}. It shows that the single master system generates
maximum traffic while cloud robotics system generates least network
traffic for the same operation. The corresponding CPU usage for the
server as well as the clients in each of these three configurations is
shown in Figure \ref{fig:cpu}. It also shows the default publishing
rate of messages at the topics for three configurations. As one can
see, a client in the single master system publishes at higher rate
(7.5 Hz) compared to that in the multi-master system (4.5 Hz) or the
cloud robotics system (4 Hz). This could be linked to the fact that
the CPU usage of the client for a single master system (SMS-C) is
lowest giving rise to higher publishing rate. A client in multi-master
system (MMS-C) and cloud robotics system (CRS-C) is required to run
additional processes to establish communication with the server which
leads to higher CPU usage and hence, lower publishing rate.  This,
however, causes more CPU usage and network usage for the server in the
single master system (SMS-S).  Overall, it appears that it is
advantageous to go for a multi-master system or cloud robotics systems
compared to a single master system as the former systems lead to lower
network traffic at a comparable CPU usage compared to the later.

\begin{figure}[!h] 
  \centering
  \includegraphics[width=3.5in]{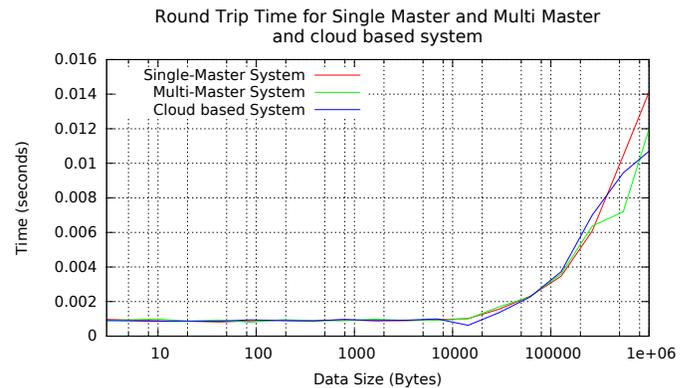}
\caption{The round trip time (RTT) for three modes of implementation:
single-master, multi-master and cloud based systems. RTT is calculated
as the time taken for message to go from a topic to another topic and
back.}
\label{fig:rtt} 
\end{figure}

We also plot the \emph{Round Trip Time} (RTT) for the three modes of
implementation. It is the time taken by a packet to go from a sender
to a receiver and come back to the sender.  In this paper, RTT is
computed as follows. A message is published at a node on one machine.
This node is subscribed by another machine, which in, turns publishes
it on another node. This new node is then subscribed by the first
time. The time difference between publishing the message on node and
receiving it at another on machine 1 is considered as the round trip
time. These two machines are located in the same place communicating
over wireless LAN. The resulting RTT for all the three configurations
is shown in Figure \ref{fig:rtt}. As expected, the round-trip time
increases monotonically with increasing data size and it's behaviour
is more or less same for all the three configurations.  Usually, the
round trip time (RTT) is computed for machines which are physically
separated by several kilometers \cite{mohanarajah2015rapyuta}.
Nevertheless, the RTT behaviour will remain more or less same as shown
in Figure \ref{fig:rtt} as the network delays between the machines
will dominate the minor differences arising out of internal processes
of each configuration.

\subsection{Experiment 2} \label{sec:expt2}

\begin{figure}[!h]
  \centering
  \begin{tabular}{c}
    \subfloat[CPU usage]{\label{fig:cpu2}\includegraphics[scale=0.6]{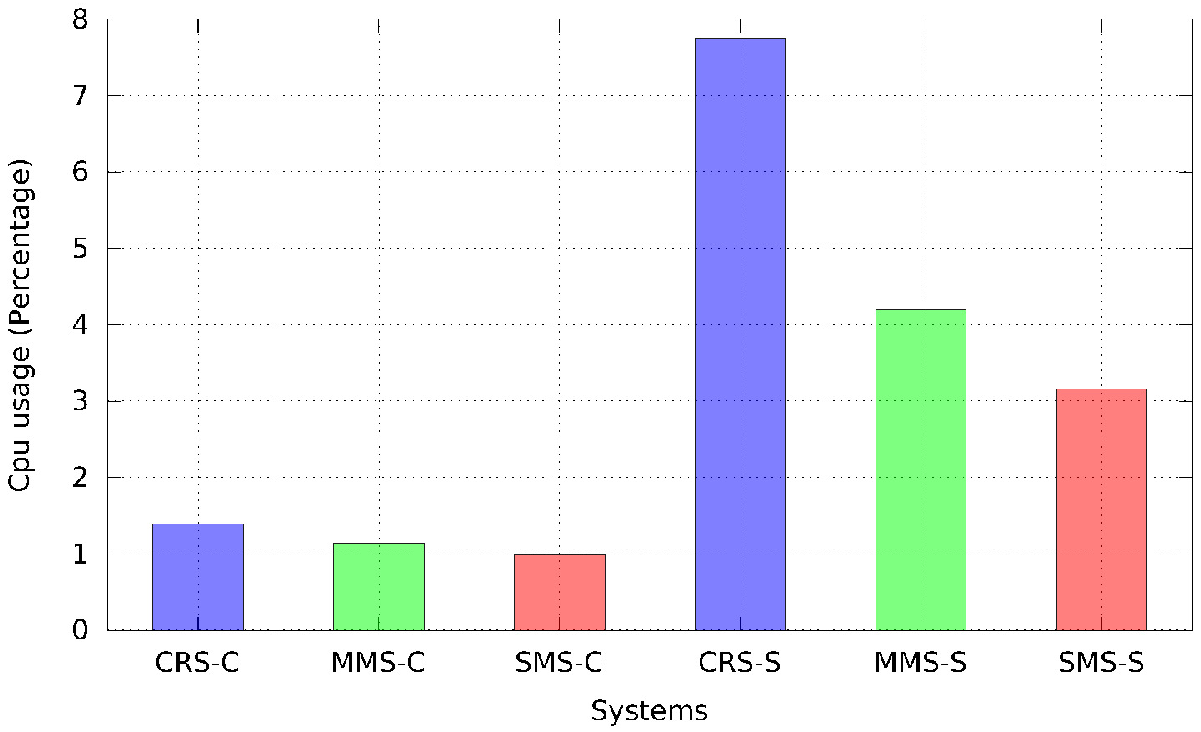}} \\
    \subfloat[Network Bandwidth Usage]{\label{fig:bw2} \includegraphics[scale=0.6]{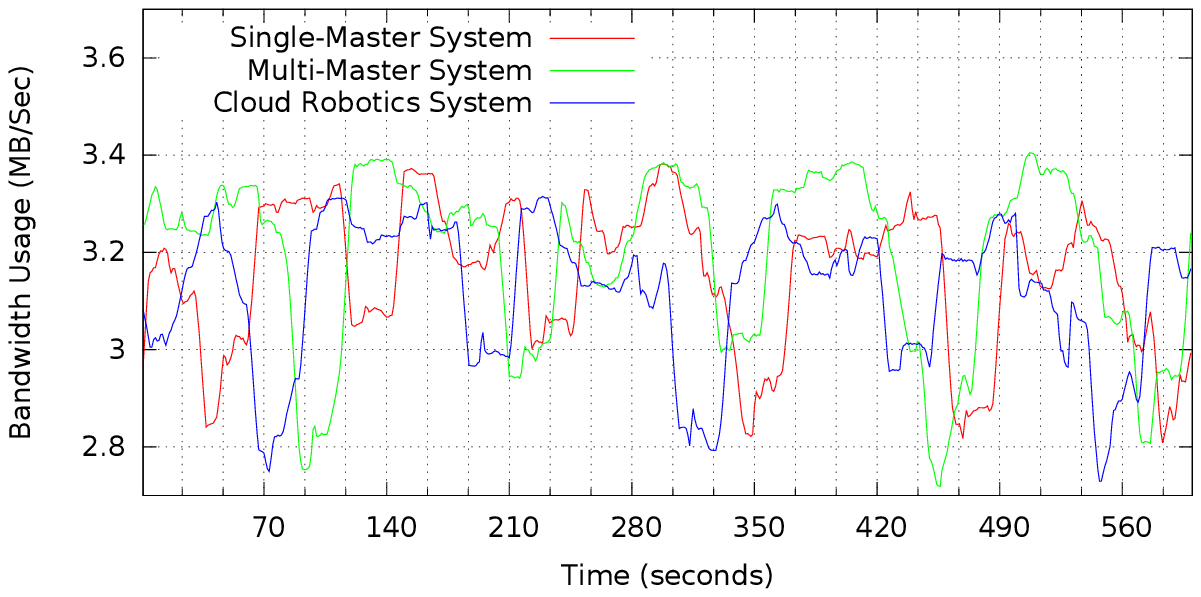}}
  \end{tabular}
  \caption{CPU and Network Usage in the second experiment.}
  \label{fig:perfcomp}
\end{figure}

In this experiment as well, two physical machines are connected to
each other through a Wireless LAN.  The experiment is further
simplified by removing the Gazebo simulator which has a high
computational as well as memory footprint. One of these machines
publish images onto a topic which is subscribed by the other machine.
The other machine simply echoes this data on a console. The second
machine subscribing to the image publishing topic is considered as the
server as it either runs a \texttt{roscore} process in the single
master mode or a Rapyuta engine in the cloud robotics mode of
operation. The relative performance of the machines is analyzed and
compared in terms of CPU usage and network bandwidth usage as shown in
Figure \ref{fig:perfcomp}.  The network usage is almost same in all
the three cases as all of them use the same publishing rate and there
are no other processes / nodes that generate additional network
traffic.  However, there is a difference in the CPU usage in these
implementations. It is highest in Cloud Robotics mode of operation
both on client as well as server side. This could be attributed to the
additional computational overhead needed for running cloud processes.
The multi-master system has the second highest CPU usage owing to the
additional computation needed for running \texttt{master\_discovery}
processes and \texttt{master\_sync} processes. Since none of these
additional processes are there in the single master mode, the CPU
usage is least in this case. These observations are in sync with our
understanding of the systems as explained in the previous sections.

\section{Limitations and Future Work} \label{sec:limit}
As summarized above, the single master or multi-master ROS systems are
not suitable for deployment of Fleet Management services as a PaaS
environment. Both these architectures implement Networked Robotics model
based on Robot-to-Robot (R2R) communication. While enabling the familiar ROS
based PaaS environment and transparent availability of sensor data
across multiple robots, the following key shortcomings or constraints
on an individual robot or a fleet of robots has to be noted: (1)
\textit{Resource Constraints} - There are resource constraints on each
robot in terms of onboard compute, memory and robot's power supply,
motion mode and working environment. Once deployed they cannot be
easily upgraded. Algorithms which require access to high dimensional
data from multiple robots requiring larger compute infrastructure will
remain constrained by the overall network of robots' compute
capacities.  (2) \textit{Communication Constraints} - higher bandwidth
usage within the R2R network of mobile robots will lead to higher
network latencies thereby deteriorating the quality of service.  (3)
\textit{Scalability constraints} - on the overall solution as number
of robots in a mobile fleet increases.  

For cloud-based PaaS systems such as Rapyuta, which implements
Robot-to-Cloud (R2C) model, the following limitations are identified
which need remediation: 

\begin{itemize}
  \item In its current form, it does not offer \emph{high
    availability} \cite{gray1991high} for Rapyuta Master taskset and
    its failure leads to collapse of the whole system. This needs
    remediation by infrastructural mechanisms in combination with
    checkpoint-restart utilities \cite{laadan2010linux}
    \cite{duell2002requirements}.  

\item Of the five key characteristics of
Cloud Services, the current implementation of Rapyuta PaaS lacks one,
namely, the \emph{elasticity}. It has a cannibalized approach for all
containers on a host to access compute, storage and network resources
on the host machine and does not offer ability to allocate and resize
these containers in the run-time as the workload changes over time.
The utilities for monitoring the resource consumption are rudimentary
and do not offer advice for migration of containers from one host to
another or resizing.  

\item In the current implementation of the cloud
platform, there are no provisions for managing communication bandwidth
to cater to different traffic situations. In practical scenarios for
fleet management, having a logical segregation of communication
bandwidth between control and data signals will improve the
responsiveness of the R2C system. This is a concern when a remote
tele-operation is required for an impaired mobile robot in a data
centric network environment. Ability to leverage Multi-Path TCP
\cite{ford2011architectural} \cite{barre2011multipath} can also
improve the transfer rates with R2C communication as it can make use
of multiple interfaces to compensate for congestion in one of the
channels. 

\item In a large warehouse of several thousand square feet area, it is
  possible that all mobile robots may not always have access to Cloud
  through the Cloud Access point. But with alternate communication
  modalities like Bluetooth, Zigbee or Wifi Direct - they may have
  connectivity to nearby robots which, in turn, may have access to the
  Cloud infrastructure. In such a scenario, a proxy-based
  \cite{hu2012cloud} compute topology will be useful where one robot
  functions as a group leader to bridge the interaction between the
  set of nearby out-of-coverage robots and the cloud. The current
  Rapyuta implementation does not provide this topology and would
  require extensive changes to enable this. However, the other
  topologies such as clone-based or peer-based models are easier to
  implement with the current implementation and may be used along the
  ROS single-master or multi-master mode to simulate proxy-based
  systems.


\item  In the current implementation of Rapyuta framework - the
  partitioning of data and compute across three options -  onboard
  compute on robot itself or robotic R2R network and/or Cloud
  execution has to be decided upfront and is usually static. Depending
  on the task with deadline, whether it is a SLAM, Navigation or
  Grasping task in warehouse, it would be useful to have a framework
  that can allocate these tasks to suitable compute resources (on edge
  / fog / cloud) in the run-time. Use of
  energy-efficient optimization algorithms \cite{vergnano2012modeling}
  \cite{dressler2005energy} for task allocation and subsequent path
  planning and coordination have to be added on the top of Rapyuta
  platform for warehouse fleet management.  
  

\end{itemize}

The directions for future work therefore include remediation of the
limitations of the Rapyuta Cloud framework and engineering the
algorithm layer for task allocation, task planning, path planning and
coordination, Grasping, Tele-operations and Collaborative SLAM in
context of Picker-to-Parts Warehouse robotics. Future work needs to
add the tier of R2R layer with adhoc network (using Multi-Master ROS)
with suitable elastic compute topology (Peer, Proxy or Clone) with R2C
Rapyuta framework.

\section{Conclusion} \label{sec:conc}

This paper presents the details of implementation of a fleet
management system for a group of autonomous mobile robots (AMR) using
three configurations: single-master, multi-master and cloud robotics
platform. The mobile robots are completely autonomous as far as their
navigation capabilities are concerned. These robots are required to
traverse the paths provided by a global planner. The global 
planner implements a basic path planning algorithm to generate paths
between the current robot locations and the desired goal locations set
by the operator, taking into account the obstacles which could be
created dynamically in run time. The whole system can be
controlled or monitored through a web-based user interface. The
details of implementation for both simulation as well as actual
experiment is provided which will be useful for students and
practicing engineers alike. These details provide an insight into the
working of each of the these modes of operation allowing us to
identify the strengths and weaknesses of each one of them. These
insights are further corroborated by analyzing parameters such as,
network usage, CPU load and round trip time.  We also identify the
critical limitations of current cloud robotics platform and provide
suggestions for improving them which forms the future direction for
our work.

\bibliographystyle{IEEEtran}
\bibliography{ref_sk}


\vspace{1cm}
\begin{minipage}[m]{\columnwidth}
  \small
\begin{wrapfigure}{l}{0.2\textwidth}
  \vspace{-0.25cm}
\centering
\includegraphics[scale=0.15]{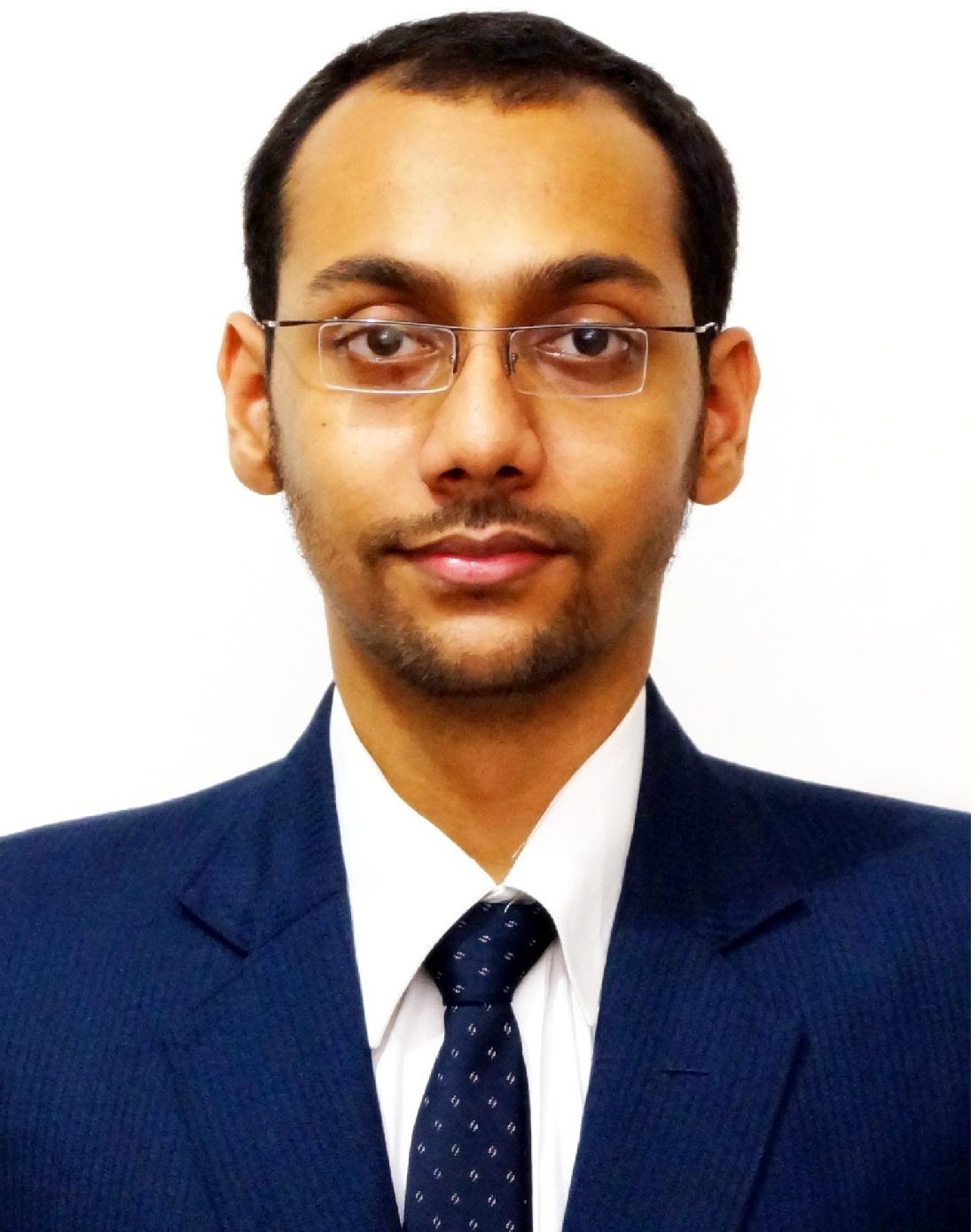}
\end{wrapfigure}
Aniruddha Singhal is working as a Researcher at Innovation Labs in
Tata Consultancy Services. He received his Bachelor's degree in
computer science from Madhav Institute of Technology, Gwalior in the
year 2014 and Master's degree in System Science from Indian Institute
of Technology Jodhpur in 2016. His current research interests include
Machine Learning, Computer Vision and Robotics
\end{minipage}

\vspace{0.5cm}

\begin{minipage}[m]{\columnwidth}
  \small
\begin{wrapfigure}{l}{0.2\textwidth}
  \vspace{-0.4cm}
\centering
\includegraphics[scale=0.12]{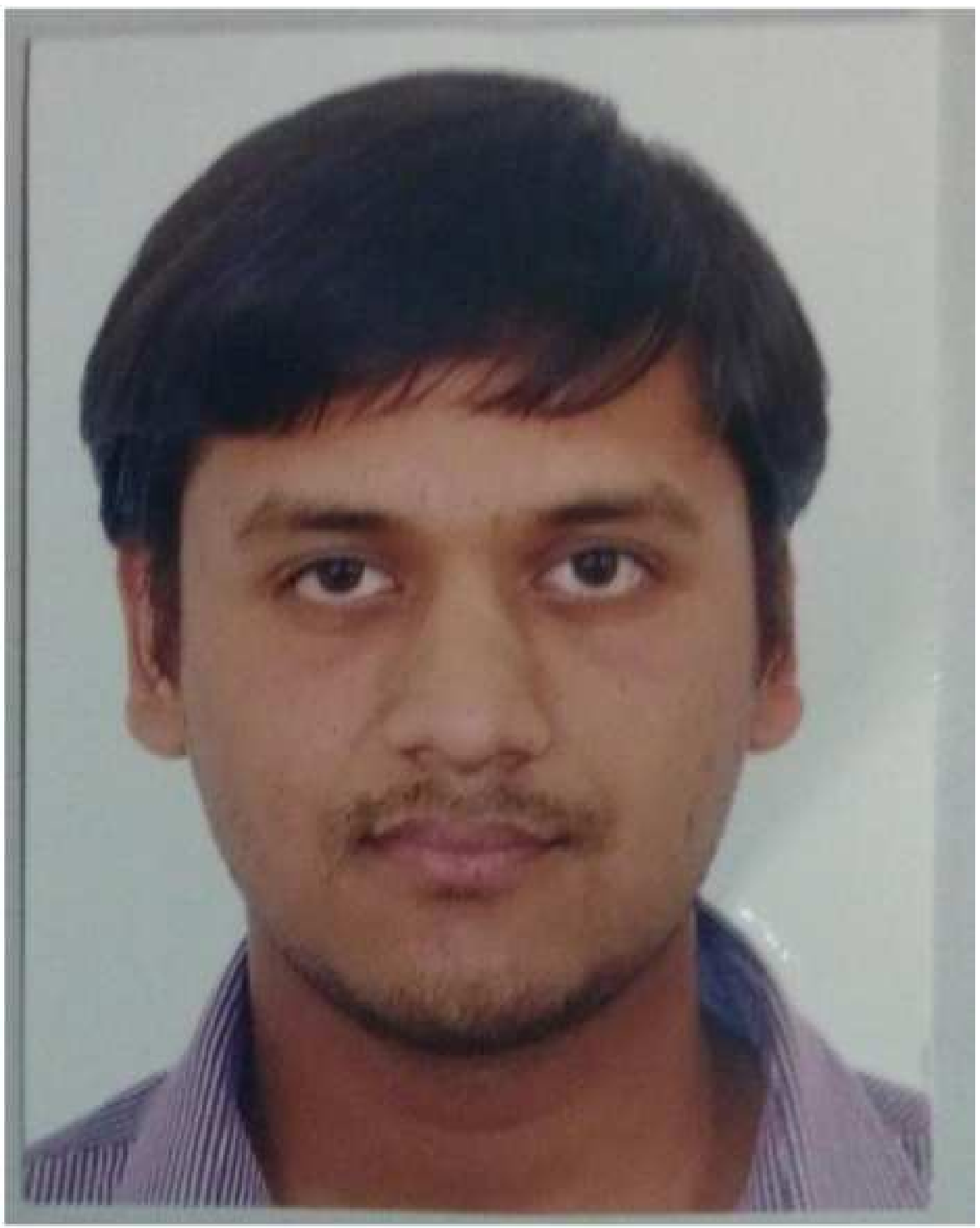}
\end{wrapfigure}
Nishant Kejriwal obtained his Bachelor's degree in Computer Science
from Indian Institute of Technology Jodhpur in 2012. Since then, he is
working as a researcher at Innovation Labs in Tata Consultancy
Services. His research interests include Machine Learning, Robotics
and Computer Vision.
\end{minipage}

\vspace{0.5cm}

\begin{minipage}[m]{\columnwidth}
  \small
\begin{wrapfigure}{l}{0.2\textwidth}
  \vspace{-0.4cm}
\centering
\includegraphics[scale=0.5]{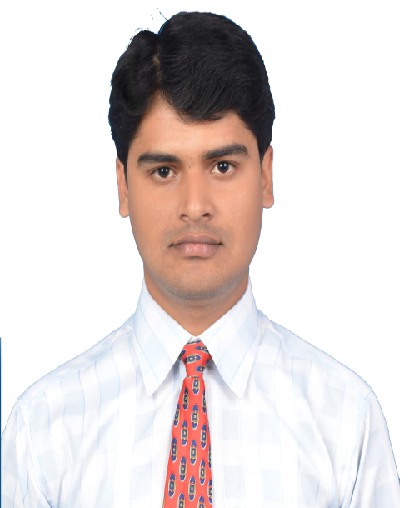}
\end{wrapfigure}
Prasun Pallav obtained his Bachelor's degree in computer science
engineering from West Bengal University of Technology in the year
2014. Since then he is working as a system engineer at Tata
Consultancy Services, New Delhi, India. His research interest includes
Linux System Programming, Robotics and Computer Vision.
\end{minipage}

\vspace{0.5cm}

\begin{minipage}[m]{\columnwidth}
  \small
\begin{wrapfigure}{l}{0.2\textwidth}
  \vspace{-0.3cm}
\centering
\includegraphics[scale=0.4]{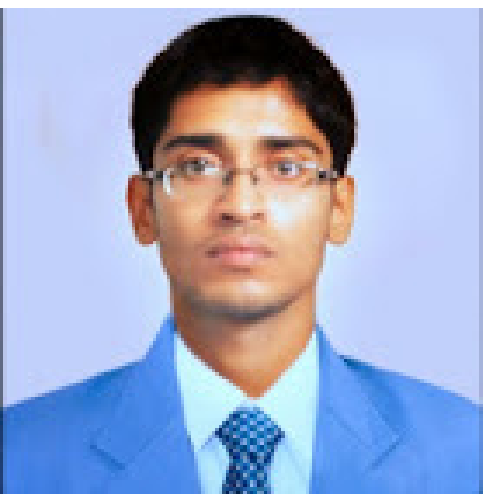}
\end{wrapfigure}
Soumyadeep Choudhury obtained his Bachelor's degree in Electronics and
Communication Engineering in the year 2015 from Academy Of Technology,
West Bengal University Of Technology. Since then, he is working as a
researcher at Innovation Labs, Tata Consultancy Services, New Delhi,
India. His research interests include Linux System Programming,
Robotics and Computer Vision.
\end{minipage}

\vspace{0.5cm}

\begin{minipage}[m]{\columnwidth}
  \small
\begin{wrapfigure}{l}{0.3\textwidth}
  \vspace{-0.3cm}
\centering
\includegraphics[scale=0.6]{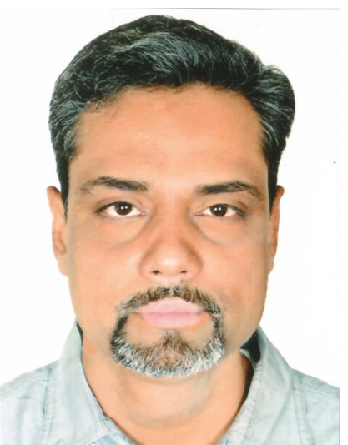}
\end{wrapfigure}
Rajesh Sinha holds a Bachelor's degree in Electrical and Electronics
Engineering from BITS Pilani and a Masters in Comparative Religion
from Dayalbagh University. He has over 20 years experience of building
engineered software and hardware solutions for Transportation,
Logistics, Government and Retail Industries and startups. He is
currently heading the Smart Machines Programme at Tata Consultancy
Services' research and innovation division. 
\end{minipage}

\vspace{0.5cm}

\begin{minipage}[m]{\columnwidth}
  \small
\begin{wrapfigure}{l}{0.2\textwidth}
  \vspace{-0.3cm}
\centering
\includegraphics[scale=0.4]{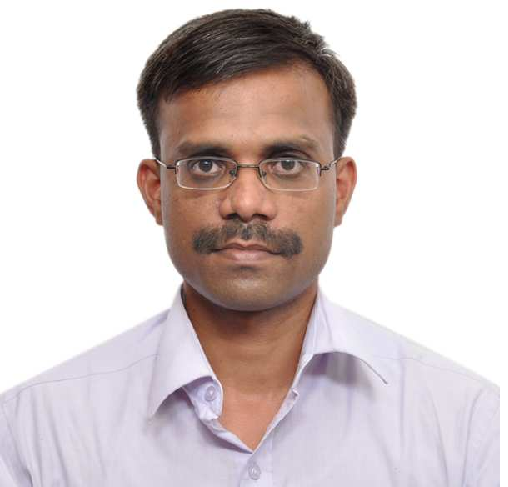}
\end{wrapfigure}
Swagat Kumar (S'08-M'13) obtained his Bachelor's degree in Electrical
Engineering from North Orissa University in the year 2001. He obtained
his Master's and PhD degree in Electrical Engineering from IIT Kanpur
in 2004 and 2009 respectively.  He was a post doctoral fellow at
Kyushu University in Japan for about a year. Then he worked as an
assistant professor at IIT Jodhpur for about 2 years before joining
TCS Research in 2012. He currently heads the robotics research group
at Tata Consultancy Services, New Delhi, India. His research interests
are in Machine Learning, Robotics and Computer Vision. He is a member
of IEEE Robotics and Automation Society. 
\end{minipage}


\end{document}